\documentclass[twoside,11pt]{article}

\usepackage{blindtext}

\usepackage{amsthm,amsmath,amssymb,amsfonts}
\usepackage{comment, xspace,diagbox,xcolor}
\usepackage[noend]{algpseudocode}
\usepackage{algorithm,algorithmicx}

\newcommand{\T}{\mathcal{T}}
\newcommand{\I}{\ensuremath{{\cal I}}\xspace}
\newcommand{\J}{\ensuremath{{\cal J}}\xspace}
\def\E{\mathbb{E}}

\def\ex{{\ensuremath{ i , \omega }}}
\newcommand{\ignore}[1]{} 

\usepackage{jmlr2e}

\newtheorem{assumption}{Assumption}

\newcommand{\new} {\color{black}}

\newcommand{\bitem}{\begin{itemize}}
\newcommand{\eitem}{\end{itemize}}
\newcommand{\benum}{\begin{enumerate}}
\newcommand{\eenum}{\end{enumerate}}
\newcommand{\bdefn}{\begin{definition}}
\newcommand{\edefn}{\end{definition}}
\newcommand{\bprop}{\begin{proposition}}
\newcommand{\eprop}{\end{proposition}}
\newcommand{\bque}{\begin{question}}
\newcommand{\eque}{\end{question}}
\newcommand{\bobsv}{\begin{observation}}
\newcommand{\eobsv}{\end{observation}}
\newcommand{\beqn}{\begin{equation}\begin{aligned}}
\newcommand{\eeqn}{\end{aligned}\end{equation}}

\newcommand{\ps}{\begin{proof}[Sketch]}
\newcommand{\brmk}{\begin{remark}}
\newcommand{\ermk}{\end{remark}}
\newcommand{\bduiqi}{\begin{aligned}}
\newcommand{\eduiqi}{\end{aligned}}
\newcommand{\bcoro}{\begin{corollary}}
\newcommand{\ecoro}{\end{corollary}}
\newcommand{\bcom}{\begin{comment}}
\newcommand{\ecom}{\end{comment}}

\newcommand{\adap}{adaptive}

\newcommand{\apxn}{approximation}

\newcommand{\avg}{average}
\newcommand{\arb}{arbitrary}

\newcommand{\alg}{algorithm}

\newcommand{\Alg}{Algorithm}

\newcommand{\assu}{assumption}

\newcommand{\bs}{\backslash}

\newcommand{\cond}{condition}

\newcommand{\contra}{contradiction}

\newcommand{\corres}{corresponding}

\newcommand{\distr}{distribution}

\newcommand{\dec}{decision}

\newcommand{\dtmnstc}{deterministic}
\newcommand{\dxs}{polynomial}

\newcommand{\emp}{empirical}
\newcommand{\expo}{exponential}
\newcommand{\elem}{element}

\newcommand{\feas}{feasible}

\newcommand{\func}{function}

\newcommand{\hypo}{hypothesis}
\newcommand{\hypos}{hypotheses}

\newcommand{\ho}{\mathbb}

\newcommand{\indep}{independent}

\newcommand{\iter}{iteration}

\newcommand{\ins}{instance}

\newcommand{\ineq}{inequality}
\newcommand{\ineqs}{inequalities}

\newcommand{\lb}{\left}
\newcommand{\rb}{\right}

\newcommand{\lar}{\leftarrow}

\newcommand{\mtx}{matrix}

\newcommand{\ow}{otherwise}
\newcommand{\omg}{\omega}
\newcommand{\Omg}{\Omega}

\newcommand{\obj}{objective}

\newcommand{\perm}{permutation}

\newcommand{\prb}{probability}

\newcommand{\parti}{particular}

\newcommand{\pmt}{parameter}

\newcommand{\param}{parameter}

\newcommand{\rv}{random variable}

\newcommand{\rar}{\rightarrow}

\newcommand{\resp}{respectively}

\newcommand{\sat}{satisfy}
\newcommand{\sats}{satisfies}
\newcommand{\satd}{satisfied}
\newcommand{\sce}{scenario}

\newcommand{\sse}{\subseteq}
\newcommand{\sps}{suppose}
\newcommand{\Sps}{Suppose}

\newcommand{\strfwd}{straightforward}

\newcommand{\submod}{submodular}

\newcommand{\unif}{uniform}

\newcommand{\unk}{unknown}

\newcommand{\xulie}{sequence}

\let\eps\varepsilon

\ShortHeadings{ODT and ASR with Noisy Outcomes}{Jia, Navidi, Nagarajan and Ravi}
\firstpageno{1}

\begin{document}

\title{Optimal Decision Tree and Adaptive Submodular Ranking with Noisy Outcomes}

\author{\name Su Jia \email sj693@cornell.edu \\
\addr Center of Data Science for Enterprise and Society\\
Cornell University
   \AND
   \name Fatemeh Navidi\email navidi@umich.edu\\
    \addr Midpoint Markets
 \AND
    \name Viswanath Nagarajan\email viswa@umich.edu \\
    \addr Industrial \& Operations Engineering\\
    University of Michigan
    \AND
    \name R. Ravi \email ravi@andrew.cmu.edu \\
    \addr Tepper School of Business\\
    Carnegie Mellon University}

\editor{N/A}

\maketitle

\begin{abstract} In pool-based active learning, the learner is given an unlabeled data set and aims to efficiently learn the unknown hypothesis by querying the labels of the data points.
This can be formulated as the classical {\em Optimal Decision Tree} (ODT) problem: Given a set of {\em tests}, a set of {\em hypotheses}, and an outcome for each pair of test and hypothesis, our objective is to find a low-cost testing procedure (i.e., {\em decision tree}) that identifies the true hypothesis.
This optimization problem has been extensively studied under the assumption that each test generates a {\em deterministic} outcome.
However, in numerous applications, for example, clinical trials, the outcomes may be uncertain, which renders the ideas in the deterministic setting invalid.
In this work, we study a fundamental variant of the ODT problem in which some test outcomes are noisy, even in the more general case where the noise is {\em persistent}, i.e., repeating a test gives the same noisy output. 
Our approximation algorithms provide guarantees that are nearly best possible and hold for the general case of a large number of noisy outcomes per test or per hypothesis where the performance degrades continuously with this number.
{\new Furthermore, most of our results  hold for a more general problem called {\em Adaptive Submodular Ranking with Noise} (ASRN).} 
We numerically evaluated our algorithms for identifying toxic chemicals and learning linear classifiers and observed that our algorithms have costs very close to the information-theoretic minimum.\footnote{A preliminary version of this paper appeared as \cite{JiaNNR19} in the {\em Proceedings of the Thirty-third Neural Information Processing Systems} (NeurIPS'19). 
This paper substantially expanded the proceedings version by (i) generalizing our results beyond decision trees to a novel
problem called {\em Adaptive Submodular Ranking with Noise} (ASRN), and (ii) extending our analysis from binary outcome space to finite outcome space.}
\end{abstract}


\section{Introduction}\label{sec:intro}

In the {\em Optimal Decision Tree} (ODT) problem, our objective is to identify an unknown true {\it hypothesis} drawn from a known {\em prior} distribution over a given set of {\em hypotheses}.
To collect information on the true hypothesis, we are also given a set of {\em tests}.
Upon selection, a test produces a binary (i.e., positive or negative) outcome that depends on the true hypothesis, and a certain cost is incurred.
Finally, we are given a binary matrix that documents the outcome of every pair of test and hypothesis. 
The goal is to find a low-cost testing procedure (i.e., decision tree) that always identifies the true hypothesis.

This fundamental problem encapsulates many real-world challenges wherein the learner aims to interactively gather information to identify the unknown ground truth.
For example, in medical diagnosis, a doctor must diagnose a patient's unknown disease by performing a low-cost sequence of medical tests, chosen from a set of available tests \citep{L85}. 
As another example, in active learning (e.g., \citealt{dasgupta2005analysis}), the learner is given a set of {\em unlabeled} data points and aims to find a correct binary classifier by efficiently querying the labels of the data points. 
% We are given a set of $m$ possible classifiers that contains the true classifier, which is drawn from a known probability distribution. 
% The goal is to identify the true classifier by querying labels using the minimum number of points in expectation.  
Other applications include entity identification in databases (\citealt{ChakaravarthyPRAM11}) and experimental design to choose the most accurate theory among competing candidates (\citealt{GolovinKR10}).

The ODT problem has been extensively studied under the assumption that each test generates a {\em deterministic} outcome. 
However, this assumption is unrealistic in many applications.
For example, in clinical trials, the results of the same medical test may vary among individuals due to genetic differences, despite the fact that they share the same underlying disease.
Similarly, in online A/B experiments, users' reactions to a particular treatment (``test'') may vary within the same user group (``hypothesis'') due to personal preferences.
% In fact, our research was motivated by a data set involving toxic chemical identification where the outcomes of many hypothesis-test pairs are stated as unknown (see Section~\ref{sec:experiments}).

Despite the considerable literature on the ODT problem, the fundamental problem of ODT with noisy outcomes is not yet adequately understood, especially from the perspective of approximation algorithms.
Previous work incorporating noise (e.g., \citealt{GolovinKR10}) was restricted to settings with very few noisy outcomes.
One of the central technical challenges in the presence of noise is that each hypothesis can potentially follow one of an {\em exponential} (in the level of uncertainty) number of trajectories. 
This leads to an unfavorable approximation ratio if the noise-free analysis is applied directly.

Against this backdrop, we embark on a comprehensive study of the fundamental problem of {\em Optimal Decision Tree with Noise} (ODTN) in full generality and design novel approximation algorithms with provable guarantees.
Essentially, we generalize the ODT problem to the setting where the test-hypothesis \mtx\ may contain some independently random entries. The positions of these entries are known but their values can only be revealed when the corresponding test is performed.
{\new We consider the {\em persistent} noise model, where repeating the same test always produces the same outcome.
% This model is (a) more general and (b) more challenging than the independent noise model, which is more common in the literature on active learning and ODT.
% In fact, to see (a), we can reduce the independent noise model to the persistent noise model by creating sufficiently many copies of each test.
% To see (b), 
Persistent noise is more challenging than independent noise, since we can no longer ``denoise'' by repeating a test many times.} 
% and reducing the problem to a deterministic one.
% However, this approach obviously fails under persistent noise.

% We consider both non-adaptive policies, where the test sequence is fixed upfront, and adaptive policies, where the test sequence is built incrementally and depends on observed test outcomes.  
% Evidently, adaptive policies perform at least as well as non-adaptive ones. 
% Indeed, there exist instances where the relative gap between the best adaptive and non-adaptive policies is very large (see, for example, \citealt{dasgupta2005analysis}).
% However, non-adaptive policies are very simple to implement, requiring minimal incremental computation, and may be preferred in time-sensitive applications. 

\begin{table}[h]
\centering
\begin{tabular}{c|ccc}
& what to choose & what is unknown & what to observe  \\ \hline
AL & unlabeled data & classifier & label  \\
ODT & test & hypothesis &   outcome \\
ASR & element & target function & response 
\end{tabular}
\label{tab:one_stone_three_birds}
\caption{One stone, three birds: Analogous terminologies in {\em active learning} (AL), {\em optimal decision tree} (ODT), and {\em adaptive submodular ranking} (ASR).}
\end{table}

Beyond the ODTN problem, our results are valid in a substantially more general setting, called {\em Adaptive Submodular Ranking with Noise} (ASRN):
Given a set of elements, we need to construct a subset of elements sequentially to cover an {\bf unknown} {\em target} function, which comes from a given family of submodular functions.
When an element is selected, we not only increase the value of the target function but also receive a random {\em response} that helps further localize the target function in the given family.
Therefore, we face a learning-versus-earning trade-off: An intelligent algorithm must consider both the coverage and the information gain when selecting the next element.
The goal is to minimize the cover time of the target function, i.e., the expected number of elements selected until the value of the target function reaches a prescribed threshold. 

The ASRN problem generalizes the ODTN problem. To see this, note that since the output must be correct with probability $1$, we need to eliminate all but one hypothesis. 
This motivates us to consider a {\em set function} for each hypothesis, whose value is proportional to
the number of other \hypos\ eliminated.
Intuitively, this function is submodular: The elimination power of the same test {\em diminishes} as we select more tests.
Our objective is to cover the submodular function of the true hypothesis, which is unknown initially but can be ``learned'' as we observe more test outcomes.
To help the reader see the connection, we list and compare analogous concepts in these problems in Table \ref{tab:one_stone_three_birds}.

In the absence of noisy outcomes, this problem has been studied in both non-adaptive \citep{azar2011ranking} and adaptive \citep{navidi2016adaptive} settings.
In addition to the ODT problem, this submodular setting captures a number of applications such as {\em Multiple-intent Search Ranking} \citep{azar2009multiple}, {\em Decision Region Determination} \citep{JavdaniCKKBS14} and {\em Correlated Knapsack Cover} \citep{navidi2016adaptive}.
Our work is the first to handle noisy outcomes in all of these applications in a unified manner.

\subsection{Contributions}
Our results can be categorized into the following four parts. 
% Throughout, we denote by $m$ the number of hypotheses.
\benum
\item {\bf Non-\adap\ Setting.} 
We first consider the non-\adap\ version of the ASRN problem, dubbed {\em Submodular Function Ranking with Noise} (SFRN). 
We obtain a polynomial-time algorithm with cost $O(\log \frac 1 \eps)$ times the optimum; see Theorem~\ref{thm:non-adp-sfrn}. 
Here, $\eps>0$ is the {\em separability} of the family of submodular functions, formally defined in Section \ref{sec:prelim}. 
This result is significant because of the following aspects.
\benum 
\item {\bf Implications for the ODTN Problem:} The above implies an $O(\log m)$-\apxn\ for the {\em non-\adap} ODTN problem where $m$ is the number of \hypos. 
This is best possible assuming ${\rm P}\ne {\rm NP}$, due to the renowned  hardness of approximation  for the Set Cover problem; see Theorem 4.4 in \citealt{feige1998threshold}.
\item {\bf Optimality:} Unless ${\rm P}= {\rm NP}$, there is no polynomial-time $o(\log \frac 1\eps)$-\apxn\ algorithm (even without noise); see Theorem 3.1 in \citealt{azar2011ranking}.
\eenum
\item {\bf Adaptive Setting with Low Noise.} We present an algorithm whose performance guarantee degrades with the noise level. 
Specifically, we introduce the notions of {\em row uncertainty} $r$ and {\em column uncertainty} $c$ (formally defined in Section~\ref{sec:few_star}), and present an $O(\min\{c,r\} + \log\frac m\eps)$-approximation algorithm for the ASRN problem where $m$ is the number of \submod\ \func s; see Theorem~\ref{thm:asrn}.
In the noiseless case, i.e., $c=r=0$, our result matches the known bound (Theorem 1 in \citealt{navidi2016adaptive}) for the special case without noise.
Our result is significant in the following respects.
\benum
\item {\bf Implications for the ODTN Problem:} By setting $\eps=\frac 1m$, we immediately obtain an $O(\min\{c,r\} + \log m)$ \apxn\ for the (adaptive) ODTN problem.
In this context, $c$ (resp. $r$) is the maximum number of noisy outcomes in each column (resp. row) of the test-hypothesis matrix.
\item {\bf Optimality Under Low Noise Level:} If the number of noisy outcomes in each row or column is $O(\log \frac m\eps)$, the approximation ratio becomes $O(\log \frac m\eps)$, which is best possible due to Theorem 4.1 in \citealt{ChakaravarthyPRAM11}.
\item {\bf Improved Approximation for the ODTN Problem:} \cite{GolovinKR10} obtained an 
%$O(\log^2 \frac1{p_{\rm min}})$-
approximation algorithm that is polynomial-time only when $c=O(\log m)$.
Our result improves the above by a logarithmic factor and is polynomial-time regardless of $c,r$.
\eenum
\item {\bf Adaptive Setting with High Noise.} So far we have focused on the case with few {\em uncertain} entries in the test-hypothesis matrix. 
Now, we consider the other extreme, where this \mtx\ has few {\it deterministic} entries. 
% Now we consider the other extreme, where the instance has high noise level.
At first sight, considering the increased level of noise, the problem appears considerably more challenging.
Surprisingly, we obtain a logarithmic approximation by combining the following components.
\benum
\item {\bf Sparsity of the Instance:} 
An ODTN instance is {\em $\alpha$-sparse} for some $\alpha\in [0,1]$ if each test has $O(m^\alpha)$ deterministic outcomes. 
The lower $\alpha$, the more challenging it is to identify the true \hypo. 
We quantify this relation by showing that the optimum is $\Omg(m^{1-\alpha})$; see Proposition \ref{lem:lower-bound}.
\item {\bf Lower Bound via Stochastic Set Cover:} As the key technical novelty, we relate the ODTN problem to the {\it Stochastic Set Cover} (SSC) problem by ``charging'' the cost to a family of SSC instances.
For each hypothesis $i$, we associate an SSC instance and show that its optimum, denoted ${\rm OPT}_{{\rm SSC} (i)}$, is a lower bound on the cost of any \alg\ 
attributed to $i$.
We then show that the optimum is at least the sum of ${\rm OPT}_{{\rm SSC} (i)}$'s, weighted by the prior probabilities.
\item {\bf A Novel Greedy Algorithm:}
Motivated by the above observation, we present a hybrid algorithm that integrates
(i) the greedy algorithm for the SSC problem and (ii) a brute-force subroutine that checks whether one of the hypotheses with the highest {\em posterior} \prb\ is the true hypothesis; see Algorithm~\ref{alg:large-noise}. 
This algorithm has a low cost since (i) the greedy SSC algorithm is an $O(\log m)$-\apxn, and (ii) the brute-force subroutine enumerates only a small number of hypotheses.
\item {\bf Approximation for $\alpha$-Sparse Instances:} Building on (b) and (c), we show that the above algorithm has cost $O(m^\alpha + \log m \cdot {\rm OPT})$ for any $\alpha$-sparse instance; see Theorem \ref{thm:sparse}. 
When $\alpha\le \frac 12$, we have ${\rm OPT}=\Omg(m^\alpha)$ due to (a), and we obtain an $O(\log m)$-\apxn.
\eenum
\item {\bf Comprehensive Numerical Experiments.} We tested our algorithms on both a synthetic and a real data set arising from toxic chemical identification. 
We compared the empirical performance guarantee of our algorithms to an information-theoretic lower bound. 
The cost of the solution returned by our non-adaptive algorithm is typically within 50\% of this lower bound, and typically within 20\% for the adaptive algorithm, demonstrating the effective practical performance of our algorithms.
\eenum

% As a final remark, although in this work we will consider \unif\ \distr\ for noisy outcomes, our results extend directly to the case where each noisy outcome has a different probability of being $\pm 1$. 
% Suppose that the probability of every noisy outcome is between $\delta$ and $1-\delta$. 
% Then our results for ASRN continue to hold, regardless of $\delta$, and the result for the many-\unk s version holds with a slightly worse $O(\frac1\delta \log m)$ approximation ratio.

\subsection{Related Work}
The ODT problem has been extensively studied for several decades; see \citealt{garey-graham,rivest-hyafil,L85,arkin1998decision,kosaraju1999optimal,adler2008approximating,chakaravarthy2009approximating,gupta2017approximation,li2020tight}. 
The state-of-the-art result is an $O(\log m)$-approximation  \citep{gupta2017approximation}, for instances with {\em arbitrary} probability distribution and costs. 
On the other hand, \cite{ChakaravarthyPRAM11} showed that ODT cannot be approximated to a factor better than $O(\log m)$ unless P=NP. 
 
The application of ODT to Bayesian active learning was formalized in \citealt{dasgupta2005analysis}. 
There are also several results on the {\em statistical complexity} of active learning; see, e.g.,  \citealt{BalcanBL06,Hanneke07,Nowak09}. 
There are two main differences from these works from ours. 
First, they focus on proving sample complexity bounds for {\em structured} hypothesis classes, such as threshold functions or linear classifiers. 
Secondly, these works primarily focus on analyzing the sample complexity, rather than comparing the cost with the optimal algorithm.
On the contrary, we consider arbitrary (finite) hypothesis classes and obtain {\em computationally efficient} policies with provable approximation bounds relative to the optimal (instance-specific) policy. 
This approach is similar to that of \citealt{dasgupta2005analysis,GuilloryB09,GolovinK11,GolovinKR10,CicaleseLS14,JavdaniCKKBS14}. 

The noisy ODT problem was studied previously in \citealt{GolovinKR10}. 
% Using a connection to adaptive submodularity.
{\new \cite{GolovinKR10} states the approximation ratio as $O(\log \frac 1 {p_{\rm min}})$, where $p_{\rm min} \le \frac 1m$ is the minimum probability of any hypothesis, but it relied on a flawed claim in \citealt{GolovinK11}; the error was pointed out by  \cite{NanS17}. 
A correct approximation ratio of %$O(\log^2 \frac 1{p_{min}})$ was established later by \cite{NanS17,GolovinK17-fixed}. Recently, it was improved to 
$O(\log \frac{n}{p_{min}})$ was obtained by \cite{esfandiari2021adaptivity}, where $n$ is the number of tests.}
\iffalse 
\cite{GolovinK11} obtained an $O(\log^2 \frac 1{p_{\rm min}})$-approximation algorithm for noisy ODT in the presence of very few noisy outcomes, where $p_{\rm min} \le \frac 1m$ is the minimum probability of any hypothesis. 
In particular, the running time of the algorithm in \citealt{GolovinKR10} is exponential in the number of noisy outcomes per hypothesis.
As noted earlier, our result improves both the running time (it is now polynomial for any number of noisy outcomes) and the approximation ratio. \fi 
We note that an $O(\log m)$ approximation ratio (still only for very sparse noise) follows from the work on the ``equivalence class determination'' problem by \cite{CicaleseLS14}. 
For this setting, our result is also an $O(\log m)$ approximation, but our algorithm is simpler. 
More importantly, ours is {\new among the first} to handle {\em any} number of noisy outcomes {\new for cost-minimization};\footnote{{\new The approach of \cite{ChenJKBSK15} can also handle a large number of noisy outcomes, but their objective is to maximize the information gained, rather than identifying the true \hypo.}} {\new other such work include \cite{gan2021greedy,gan2021toward}.} 

Other variants of noisy ODT have also been considered, where the goal is to identify the correct hypothesis with at least some target probability \citep{NaghshvarJC12,BellalaBS11,ChenH017}. 
\cite{ChenH017} provided a bi-criteria approximation in which the algorithm has a higher error probability than the optimal policy. 
Our setting is different because we require {\bf zero} probability of error.  

Many results for ODT (including some of ours) rely on certain submodularity properties. 
We briefly survey some background results. 
In the basic {\em Submodular Cover} problem, we are given a set of elements and a submodular function $f$. 
The goal is to use the minimal number of elements to increase the value of $f$ to reach a certain threshold.
\cite{wolsey1982analysis} first considered this problem and proved that the natural greedy algorithm is a $(1 + \ln \frac{1}{\eps})$-approximation, where $\eps$ is the minimal positive marginal increment of the function. 
As a natural generalization, in the Submodular Function Ranking problem we are given {\it multiple} submodular functions and aim to {\it sequentially} select elements so as to minimize the total cover time of these functions.
\cite{azar2011ranking} proposed a best-possible $O(\log \frac{1}{\epsilon})$-approximation algorithm for this problem, and \cite{im2016minimum} extended this result to also handle arbitrary costs. 
More recently, \cite{navidi2016adaptive} studied an adaptive version of the submodular ranking problem and presented a best-possible $O(\log \frac m \eps)$-approximation where $m$ is the number of functions.

Finally, we note that there is also work considering the {\em worst-case} (instead of average case) cost in ODT and active learning; see, e.g., \citealt{Moshkov10,SaettlerLC17,GuilloryB10,GuilloryB11}. 
These results are incomparable to ours because we are interested in the average cost.
Moreover, the analysis of average cost is, in general, more intricate than that of the worst-case cost.

\section{Preliminaries}\label{sec:prelim}

% In this section, we formally define the problem and formulate useful concepts that will be used throughout.

\def\op{\pi}
In the problem of {\em Optimal Decision Tree with Noise} (ODTN), we are given a set of $m$ possible {\it hypotheses} with a {\it prior} probability distribution $\{\pi_i\}_{i=1}^m$, from which an unknown true hypothesis $\bar{i}$ is drawn. 
There is also a set $\T$ of $n$ binary {\it tests}. 
Each test $T\in \T$ is a mapping $T:[m]\rar \{+1,-1,\star\}$. {\new  Equivalently, a test is a three-way partition $T^+\cup T^-\cup T^*$ of $[m]$, where  $T^o = \{h\in [m]: T(h) = o\}$ for each $o\in \{+,-,\star\}$.}
When this test is performed, we will observe an outcome $T(\bar i)$ if $T(\bar i)\neq \star$, and observe $+,-$ with probability $\frac 12$ if $T(\bar{i} )= \star$.
We assume that the random outcomes are {\bf independent} conditioned on the true \hypo.\footnote{{\new The \indep\ noise assumption is somewhat strong, as it disallows correlation between the test outcomes, conditional on $\bar i$. 
This \assu\ is commonly used in previous works; see, e.g., Section 3.1 of \citealt{ChenJKBSK15}.}}

%where the outcome of test $T$ is (a) positive if $\bar{i}\in T^+$, (b) negative if $\bar{i}\in T^-$, and (c) positive or negative with probability $\frac 12$ if $\bar{i}\in T^*$.

Alternatively, we can view an instance as a matrix $M \in \{+1,-1,\star\}^{n\times m}$, where each $\star$-entry is independently drawn from $+1$ and $-1$ uniformly. 
We emphasize that we only know the positions of the $\star$ entries but not their realized binary values, which can only be revealed when the corresponding test is selected.

We aim to identify $\bar i$ by iteratively eliminating \hypos. 
\Sps\ we select a test $T$ and observe an outcome $O\in \{\pm 1\}$. 
Then, we can rule out the \hypos\ $i\in [m]$ with $T(i) = -O$. 
We emphasize that we can not rule out \hypos\ $h$ with $T(h) = \star$. 
In fact, if $h$ is the true \hypo, then there is still non-zero \prb\ that we will observe $O$ when $T$ is selected.

We consider the {\it persistent} noise model. That is, repeating a test $T$ with $\bar{i}\in T^*$ always produces the same outcome. 
This model is (a) more general and (b) more challenging than the independent noise model, which is more common in the literature on active learning and ODT.
In fact, to see (a), we can reduce the independent noise model to the persistent noise model by creating sufficiently many copies of each test.
To see (b), note that in the independent noise model, we can ``denoise'' by repeating a test many times and reducing the problem to a deterministic one.
However, this approach fails under persistent noise.

We require that the output be correct with \prb\ $1$. 
To ensure that this is \feas, we assume that the true hypothesis $\bar{i}$ can be uniquely identified by performing all tests, regardless of the outcomes of $\star$-tests, i.e., tests $T$ where $T(\bar i) = \star$.

\begin{assumption}[Identifiability]\label{assu:iden} For any \hypos\ $i,j\in [m]$, there exists a test $T$ such that $T(i) \neq T(j)$ and $T(i), T(j)\in \{+1,-1\}$.
\end{assumption}

Many of our results still hold (with possibly weaker guarantees) without the identifiability assumption; see Appendix \ref{app:non-id}.
Our goal is to minimize the expected number of tests performed.
We formally define the cost when we introduce the more general problem of ASRN in Section \ref{sec:SFRN_ASRN}.

\section{Submodular Function Ranking and Its Variants}\label{sec:SFRN_ASRN}

Many of our results for the ODTN problem are obtained as corollaries of a more general problem, {\em Adaptive Submodular Ranking with Noise} (ASRN). 
To define this problem, we first review the basic versions. 
In Section \ref{subsec:SFR}, we introduce the {\em Submodular Function Ranking} (SFR) problem \citep{azar2011ranking}, where elements are selected {\em non-\adap ly} to cover a family of submodular \func s.
Then, in Section \ref{subsec:asr}, we review the adaptive version of SFR, called the {\em Adaptive Submodular Ranking} (ASR) problem \citep{navidi2016adaptive}, where the elements are selected to cover an unknown target submodular function, adaptively based on observed information on the target function.
Finally, in Section \ref{subsec:asr-noise}, we dive into full generality by introducing the problem of {\em Adaptive Submodular Ranking with Noise} (ASRN), which generalizes both the ASR and ODTN problems.

\subsection{Submodular Function Ranking, Noiseless Case}\label{subsec:SFR}
Let us begin with the simplest setting and gradually add components in the next two subsections.
\cite{azar2011ranking} introduced the following {\em Submodular Function Ranking} (SFR) problem. 
We are given a {\em ground set} of {\it elements} $[n]:=\{1,...,n\}$ and a collection of monotone submodular functions $\{f_1,...,f_m\}$ where $f_i : 2^{[n]}\rightarrow [0,1]$ satisfies $f_i(\emptyset)=0$ and $f_i([n])=1$ for all $i\in [m]$. 
It is without loss of generality (w.l.o.g.) to assume that the range is $[0,1]$, since any bounded function can be normalized to take values in $[0,1]$.
Each $i\in [m]$ is called a \textit{scenario}.
An \unk\ \textit{target} scenario is drawn from a known distribution $\{\pi_i\}$ over $[m]$. 

Note that in this problem, we are not able to ``learn'' the target \func\ based on any observable information.
Therefore, a decision rule can be formulated as a permutation of elements.
For a fixed permutation $\sigma$, we define the {\it cover time} of a scenario $i$ as the first time $f_i$ reaches the value $1$ if we select elements one by one according to $\sigma$.
The objective in the SFR problem is to find a permutation $\sigma$ of $[n]$ with minimal expected cover time.

\bdefn [Cover Time and Cost]
Let $\sigma= (\sigma(1),\dots,\sigma(n))$ be any \perm\ of the \elem s and $i\in [m]$ be a \sce. Then, the {\bf cover time} is defined as 
\[C(i, \sigma) := \min\lb\{t \ \vert f_i(\{\sigma(1),...,\sigma(t)\})= 1\rb\}.\] 
The {\bf cost} of $\sigma$ is ${\rm Cost}(\sigma) :=\sum_{i\in [m]} \pi_i \cdot C(i,\sigma).$
\edefn
 
\cite{azar2011ranking} proposed a greedy algorithm that constructs a permutation of elements by iteratively selecting the next element with the highest {\em score}.
This score assigns higher priority to those scenarios close to being covered. Specifically, the weight of each scenario is inversely proportional to the distance from $1$ and the current value of $f_i$.
We will formally state this algorithm in the form of pseudo-code in \Alg\ \ref{alg:non-adp} after we introduced the noisy variant in the next subsection.

This algorithm has the best possible approximation ratio in terms of {\em separability} \param\ $\eps>0$, defined as the minimum positive marginal increment of any function. 

\bdefn[Separability]
Given a family of non-decreasing functions $\mathcal F$, its {\bf separability} is defined as 
\[\eps:= \min\{f_i(S\cup \{e\})-f_i(S)>0\ |\ \forall S\subseteq [n], i\in [m], e\in [n]\}.\]
\edefn

\cite{azar2011ranking} showed the following in their Theorem 2.1.
\begin{theorem}[\citealt{azar2011ranking}]
\label{thm:sfr}
There is a polynomial-time algorithm whose cost is $O(\log \frac 1 \eps)$ times the optimum for any SFR instance with separability \pmt\ $\eps>0$.
\end{theorem}

% A natural idea is to reduce SFRN to SFR as follows: the elements set is still $[n]$, for each expanded scenario $(i,\omg)\in H$ we associate a submodular function $f_{i,\omg}$ with weight $\pi_{i,\omg} = \Pr(\omg|i)\cdot \pi_i$, where $f_{i,\omg}$ is defined as in section~\ref{sec:exp_hypo}.
% To see the equivalence, note that a solution to SFRN is also a permutation $\sigma$ of elements $[n]$, moreover, the objective of SFRN is the expected number equals
% $$Obj_{SFRN}(\sigma):=\sum_{(i,\omg)\in H} \pi_{i,\omg} \cdot C(i,\sigma|\omg)
% = \sum_{(i,\omg)\in H} \pi_{i,\omg} \cdot C(f_{i,\omg},\sigma) =:Obj_{SFR}(\sigma),$$
% where $\sigma$ is any \perm\ (policy) for the SFR \ins. 
% It now follows that this SFR instance is equivalent to the original SFRN \ins.

\subsection{Adaptive Submodular Ranking, Noiseless Case}\label{subsec:asr}
An instance in the {\em Adaptive Submodular Ranking} (ASR) problem is slightly more involved than in the SFR problem in the following {\bf two} ways. 
First, for each scenario $i\in [m]$, there is a known {\em response function} $r_i:[n] \rar \Omg$ where $\Omg$ is a finite set of {\em responses} (or {\em outcomes}, which we use interchangeably).
If $i$ is the true scenario and an element $e\in [n]$ is selected, then a response $\omg = r_i(e) \in \Omg$ is generated and thus any scenario $j$ with $r_j(e)\ne \omg$ can be eliminated. 
Second, the domain of each submodular function is expanded to incorporate the variability of the responses: The domain for each \submod\ function is $2^{[n]}$ in an SFR instance, and is instead $2^{[n]\times \Omg}$ in an ASR instance.
The SFR problem can be cast as a special case of the ASR problem: The reduction is immediate by setting the response set $\Omg$ to a singleton.
%The goal is to adaptively find a sequence of elements that minimizes the expected cost to cover a random scenario $\bar i$ drawn from $\{w_i\}$.

An {\em adaptive policy} (or {\em decision tree}) constructs a sequence of elements incrementally and adaptively, based on the responses of the previous elements. 
A policy is simply a function that maps the current {\em state}, i.e., elements selected so far and their responses, to an element that will be selected next. 
We formalize this concept below.

\bdefn[Adaptive Policy] 
The {\bf state} is a tuple $(E,O)$ where $E\sse [n]$ and $O\in \Omega^E$.
An {\bf adaptive policy} is a mapping $\Phi:2^{[n]\times \Omega}\rightarrow [n]$.
\edefn 

Similarly to the non-adaptive setting, we aim to minimize the expected cover time.
Let $\Phi$ be an adaptive policy.
Observe that, conditional on any true scenario $i\in [n]$, the sequence of elements selected is uniquely determined by $\Phi$. 
In fact, this sequence can be specified inductively and explicitly as follows.
Suppose that elements $e_1,\dots, e_k$ have been selected in the first $k$ iterations. 
Then, the responses are $r_i(e_1),\dots, r_i(e_k)$. 
By the definition of $\Phi,$ the next element selected would be $e_{k+1} := \Phi(\{(e_t, r_i(e_t)): t=1,\dots,k\})$.
We denote this sequence by $\sigma_{i,\Phi}$ and define the cover time as follows.

\bdefn[Cover Time, Adaptive Setting]\label{defn:cover_time}
Let $\Phi$ be a policy and $i\in [m]$ be a \sce. 
Suppose $e_1,\dots, e_n$ is the (deterministic) sequence of elements selected by $\Phi$ if $i$ is the true scenario.
The {\bf cover time} of $i$ is then defined as \[C(i, \Phi) := \min\{k \mid f_i(\{(e_t, r_i(e_t)): t\in [k]\})= 1\}.\]
The {\bf expected cover time} is ${\rm ECT}(\Phi) := \sum_{i\in [m]} \pi_i \cdot C(i,\Phi).$
\edefn

The \obj\ of the ASR problem is to find an adaptive policy $\Phi$ with minimal expected cover time.
\cite{navidi2016adaptive} showed a best-possible approximation algorithm (see their Theorem 1) that we will apply in Section~\ref{sec:few_star}.

\begin{theorem}[\citealt{navidi2016adaptive}] \label{thm:asr} There is a polynomial-time algorithm whose cost is $O(\log \frac{m}{\epsilon})$ times the optimum for any ASR instance with separability \pmt\ $\eps>0$.
\end{theorem}
Note that the ASR problem is a generalization of the (noiseless) ODT problem. 
In fact, for any \hypo\ $i$ in the ODT problem, we can define a submodular function $f_i$ that maps a subset of tests to the number of other \hypos\ eliminated by these tests, if $i$ is the true \hypo.

Analogously, we next introduce a noisy version of the ASR problem and show that it generalizes the ODTN problem.

\subsection{Adaptive Submodular Ranking with Noise}\label{subsec:asr-noise}
We now formally define the problem of {\em Adaptive Submodular Ranking with Noise} (ASRN).
An ASRN instance is almost identical to that of an ASR \ins: We are given a set of $n$ elements and a set of $m$ scenarios. 
Each scenario $i\in [m]$ is associated with a known submodular function $f_i: 2^{[n]\times \Omg}\rar [0,1]$. 
We are also given a known prior distribution $(\pi_i)$ over the scenarios.

The {\bf only} difference lies in the {\em response \func}:
For each scenario $i$, the response function $r_i:[n] \rar \Omg\cup\{\star\}$ can take a special value $\star$. 
Suppose $i\in [m]$ is the true hypothesis and $r_i(e) =\star$, then the response will be {\bf \indep ly} drawn from a known \distr\ on $\Omg$. 
For simplicity, we will consider uniform \distr, although our results extend to arbitrary distributions.

Although the responses are random, we can still use them to eliminate scenarios.
To see this, take $\Omg = \{\pm 1\}$. Suppose $i\in [m]$ is the true \hypo\ and $r_i(e)=\star$ for some element $e\in [n]$. 
When $e$ is selected, we observe $+1, -1$ with probability $1/2$. 
If $+1$ is observed, then we eliminate every scenario $j$ with $r_j(e) = -1$. 
Similarly, if $-1$ is observed, then we eliminate every scenario $j$ with $r_j(e) = +1$. 
In other words, a random outcome helps eliminate a {\em random} subset of scenarios.

As a key technique challenge, unlike in the deterministic case, a scenario $i\in [m]$ may follow {\it multiple} (more precisely, exponentially many in the number of $\star$'s) paths in the decision tree corresponding to policy $\Phi$. 
To formally define the cover time, we observe that, conditioned on the realized responses of {\em all} elements, the policy selects a {\em deterministic} \xulie\ of elements. 
To formalize this idea, we need the following notion of {\em consistent} vectors. 

\bdefn[Consistency of Response Vectors] A vector $\omg=(\omg_e)_{e\in [n]} \in \Omg^n$ is {\bf consistent} with a scenario $i\in [m]$ if for any element $e\in [n]$ with $r_i(e)\neq \star$, it holds that $r_i(e) = \omg_e$.
\edefn

% Next, we define the cover time.
% The definitions of non-adaptive and adaptive policies are identical to the noiseless setting.
% The cover time of a scenario $i$ can also be analogously defined as the first time that $f_i$ is covered by the element-response pairs observed so far; see Definition \ref{defn:cover_time}.   
% However, unlike in the noiseless setting, now the cover time can be random due to the randomness of the response set. 
% The goal is to cover the target scenario using the minimum expected number of elements.

Using terminologies from probability theory, conditioned on the event that the true scenario is $i$, we can view $\Omg^n$ as the ground set (for the probability space) and $\omg$ as a ``sample path''. 
This probability space is equipped with a uniform {\new probability mass function $(p_{\omg|i})$} over all sample paths $\omg$ {\em consistent} with $i$.
Next, we define {\em conditional} cover time as a \rv\ (i.e., a function defined on $\Omg^n$) that maps each sample path $\omg$ to the cover time conditioned on $\omg$.

\bdefn[Conditional Cover Time]
Let $\Phi$ be an adaptive policy. Let $i\in [m]$ be a scenario and 
$\omg\in \Omg^n$ be a consistent vector.
We denote by $\sigma_{i,\omg}= (\sigma_{i,\omg}(1),\dots, \sigma_{i,\omg}(n))$ the unique \xulie\ of elements selected by $\Phi$ if $i$ is the true scenario and the responses are given by $\omg$. 
We write $\sigma:=\sigma_{i,\omg}$ as a shorthand and define the {\bf conditional cover time} as
\[C(i,\Phi|\omg):= \min\{k\ |\ f_i(\{ (e_{\sigma(1)}, \omg_{\sigma(1)}),\dots,(e_{\sigma(k)}, \omg_{\sigma(k)})\}) =1\}.\]
\edefn

To define the cost of a policy, we take the expectation over (i) all \sce s and (ii) all sample paths, conditional on a \sce.
% So far we have seen that the cover time for each sample path $\omg$ is uniquely defined. 
% Therefore, we can define the cover time as a random variable,  that maps each sample path $\omg\in \Omg^n$ to the \corres\ cover time of scenario $i$.
% We formalize this idea as follows.

\bdefn[Cost of a Policy]
Let $\Phi$ be a policy and $\omg\in \Omg^n$.
Let $p_{\omg|i}$ be the probability mass of $\omg$ when $i$ is the true \hypo, and define the {\bf expected cover time} of $i$ as
$\mathrm{ECT}(i,\Phi) := \sum_{\omg\in \Omg^n} p_{\omg|i} \cdot C(i, \Phi|\omg).$ 
The {\bf cost} of $\Phi$ is defined as \[{\rm Cost}(\Phi) := \sum_{i\in [m]} \pi_i \cdot \mathrm{ECT}(i,\Phi).\]
% The {\bf cover time} of $i$ is $C(i,\Phi) = \min \{k: f_i(\{(E_1,O_1),\dots, (E_k,O_k)\})=1\}$
\edefn

To ensure the existence of a policy with {\em finite} cost, we need an assumption analogous to the identifiability assumption for the ODTN problem (Assumption \ref{assu:iden}). 
We assume that for each scenario $i\in [m]$, the function $f_i$ can be covered w.p. $1$ if we select all elements.

\begin{assumption}[Feasibility of Coverage]
For any scenario $i\in [m]$ and {\bf any} $\omega\in \Omega^n$ consistent with $i$, we have $f_i(\{(e, \omega_e) : e\in [n]\})=1$.
\end{assumption}

An important special case is where $\Omg$ is a singleton set.
{\new In this case, adaptivity does not
provide any additional advantage because we never observe anything informative.
This setting is equivalent to the SFR problem.}

% We completely settle the SFRN problem in Section \ref{sec:nonadap}, thus setting the stage for the ASRN problem in Section \ref{sec:few_star}. 

% \su{Work on this parag.}
% \rmk{Finally, we emphasize several key aspects that the ASRN problem differs from the {\em Adaptive Submodularity} (AS) framework of \cite{GolovinK11}.
% First, in the AS problem the target function is known, where in our ASRN problem it is unknown and has to be inferred using the responses observed.}

\begin{figure}
    \centering    \includegraphics[width=75mm,scale=0.5]{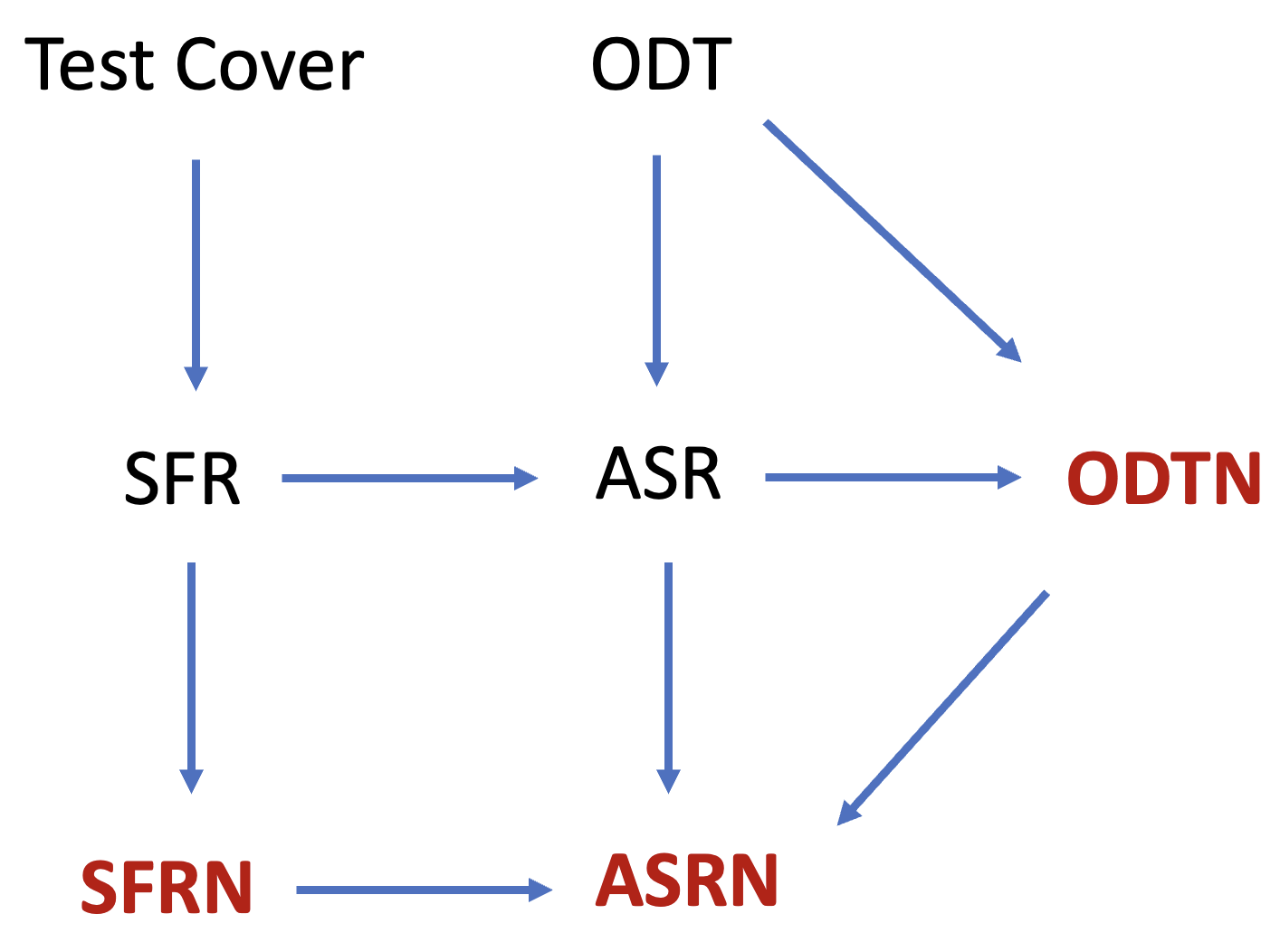}
    \caption{Connections between related problems: Edges represent (direct) reductions between problems.
    The {\em test cover} problem \citep{de2003approximation} , which was not mentioned so far, is essentially a non-adaptive version of the ODT problem, and hence can be reduced to the SFR problem. 
    We highlight the new problems introduced in this work in red color.}
    \label{fig:DAG}
\end{figure}

\subsection{Connection to the ODTN Problem} 
We illustrate the connections between the problems in Figure \ref{fig:DAG}.
We observe that the ODTN problem can be reduced to the ASRN problem. 
Let us view the tests and  \hypos\ in the ODTN problem as elements and scenarios \resp\ in the ASRN problem.
For any \hypo\ $i\in [m]$, define its response function $r_i(T) = T(i)\in\Omg\cup \{\star\}$. 
Furthermore, for any $i\in [m]$ and any $S\sse \T\times \{\pm 1\}$, we define a submodular function
\[f_i(S) = \frac 1{m-1} \lb| \bigcup_{T: (T,+1)\in S} T^- \,\bigcup \,\bigcup_{T: (T,-1)\in S} T^+\rb|,\]
where we recall that each test is a three-way partition $(T^+, T^-, T^*)$ of $[m]$.
In words, $f_i(S)$ is the fraction of hypotheses (other than $i$) that are incompatible with at least one outcome in $S$. 

It is easy to see that each function $f_i: 2^{[n]\times \Omg} \rightarrow [0,1]$ is monotone and submodular. 
Furthermore, the separability parameter $\eps =\frac{1}{m-1}$. 
More importantly, we observe that $i$ is identified if and only if the function $f_i$ has value $1$.
The reduction follows by combining the above observations.
 
\subsection{Expanded Scenario Set}\label{sec:exp_hypo}
For both non-adaptive and adaptive settings, given an ASRN instance $\I$, we will consider an equivalent ASR instance $\J$.
Thus, we can apply known algorithms for the ASR problem to the ASRN setting, and immediately bound the approximation ratio.

We emphasize that this does {\bf not} suggest that our results are mere straightforward extensions of the known results for the ASR problem.
In fact, the instance $\J$ is {\em exponentially} large compared to the original instance $\I$, and therefore it is highly non-trivial to find an efficient implementation of the ASR-based algorithm. 
In fact, most of Section \ref{sec:nonadap} and Section \ref{sec:few_star} is dedicated to elucidating our efficient implementation.

In this subsection, we focus on explaining how to define the equivalent ASR \ins.
Let $\cal I$ be a given ASRN instance with scenarios $[m]$, submodular functions
$\{f_i(\cdot)\}$ and response functions $\{r_i(\cdot)\}$. 
In the ASR instance $\cal J = \cal J(\cal I)$, each scenario in the original instance is divided into an exponential number of {\em expanded scenarios}, each corresponding to a sample path.

\bdefn[Expanded Scenarios]\label{def:expanded_sce} For each $i\in [m]$, denote  
\begin{align*} \Omega(i) &:=\{\omega\in \Omega^n: \omg\  {\rm is\ consistent\ with}\ i\} \\
&= \{\omega\in \Omega^n \,:\, \omega_e = r_i(e) {\rm \ for\ all\ } e\in [n] {\rm \ with\ } r_i(e)\ne \star\}.
\end{align*} 
An {\bf expanded scenario} is a tuple $(i, \omega)$ where $\omega\in \Omega(i)$.
% corresponds to the original scenario $i\in [m]$ when the outcome of each element $e$ is $\omega_e$. 
Furthermore, we denote $H_i:=\{(i,\omega) \,:\, \omega \in \Omega(i)\}$ and $H:=\bigcup_{i=1}^m H_i$.
\edefn
% Note that an expanded scenario also fixes all noisy outcomes.  

To define the prior \distr\ in the new \ins, for a fixed scenario $i$, consider $c_i:=|\{e\in [n] : r_i(e)=\star\}|$. 
Since the response of any $\star$-element for $i$ is \unif ly drawn from $\Omg$, each of these $|\Omega|^{c_i}$ possible expanded scenarios occurs with the same probability $\pi_{i,\omega} = \pi_i/|\Omega|^{c_i}$.
\footnote{{\new For a general noise \distr, we can redefine  $\pi_{i,\omg} = p_{\omg\mid i}\cdot \pi_i$, where we recall that  $p_{\omg\mid i}$ is the \prb\ mass \func\ of the sample path $\omg$ conditional on \hypo\ $i$.}}
To complete the reduction, for each $(i,\omega)\in H$, we define the response function $r_{i,\omega} : [n]\rightarrow \Omega$ where
$$r_{i,\omega}(e) = \omega_e,\qquad \forall e\in [n],$$
and the submodular function $f_{i,\omega} : 2^{[n]}\rightarrow [0,1]$ where \[f_{i,\omega}(S) = f_i\lb(\{(e,\omega_e\rb)\}_{e\in S}),\qquad \forall S\sse [n].\]
By this definition, since $f_i$ is monotone and submodular on $[n]\times \Omega$, the function $f_\ex$ is also monotone and submodular on $[n]$. 
% We will also work with the (noiseless) 
% ASR instance on the expanded scenarios with response functions $r_{i,\omega}$ and submodular functions $f_{i,\omega}$.
We will formally show the following in Appendix~\ref{apdx:reduction}.

\begin{proposition}[Reduction to the Noiseless Setting]\label{prop:equivlence}
The ASRN instance $\cal I$ is equivalent to the ASR instance $\cal J$. 
% Formally, the expected cost of any policy on any ASRN instance $\cal I$ is equal to the expected cost of induced policy ASR instance $\cal J$.
\end{proposition}

We reiterate that the number of expanded scenarios can be exponential in the number of uncertain entries, and therefore we cannot directly apply the existing algorithms for the ASR problem. 
We explain how to circumvent this issue in  Section \ref{sec:nonadap} and Section 
\ref{sec:few_star}.

\section{The Non-\adap\ ASRN Problem}\label{sec:nonadap}
This main result in this section is an $O(\log \frac 1\eps)$-approximation for the SFRN problem, where we recall that $\eps>0$ is the minimal marginal increment of any \submod\ function in the given family.
As a corollary, we obtain an $O(\log \frac 1m)$-approximation for the non-adaptive ODTN problem where $m$ is the number of \hypos.

% \rmk{ By Proposition \ref{prop:equivlence}, the SFRN problem is \equi\ to the SFR problem on the expanded scenarios.
% However, as the main technical challenge, there are an exponential number of expanded scenarios, and hence a direct application of the algorithm in 
% Theorem \ref{thm:sfr} leads to an exponential running time.  
% To circumvent this, we devise an efficient randomized subroutine that finds an element with approximately the highest greedy score. 
% Specifically, for each \elem\ we randomly sample a polynomial number of sample paths 
% The  estimation guarantees a (multiplicative) constant factor
% We formalize these ideas in the rest of this section.}

\subsection{The Greedy Score} \label{subsec:greedy_score}
\cite{azar2011ranking} proposed a greedy algorithm for the SFR problem.
We rephrase this algorithm in the context of expanded scenarios.
\Sps\ we have selected a set $E$ of \elem s.
We then select the next \elem\ to be the one with the highest {\em score}, which measures the additional coverage it provides when selected. 

\bdefn[Non-\adap\ Greedy Score]
Let $E\sse [n]$ be a subset of \elem s. 
Then, for each $e\in [n]\setminus E$, we define
\begin{equation}\begin{aligned}
\Delta_E(i,\omg,e) := 
\begin{cases}
\frac{f_{i,\omg}(\{e\}\cup E) - f_{i,\omg}(E)}{1-f_{i,\omg}(E)}, &\text{if}\ f_{i,\omg}(E)<1,\\
0, &\text{otherwise.}
\end{cases}
\end{aligned}\end{equation}
Furthermore, we define the {\bf greedy score} as
\begin{equation} \label{eqn:greedy_rule}
G_E(e) := \sum_{(i,\omg)\in H} \pi_{i,\omg} \cdot \Delta_E(i,\omg;e),    
\end{equation}
\edefn
% where $\Delta_E(b,h;T)$ is the ``gain'' of \elem $e$ for the hypothesis-copy $(b,h)$. 
Let us understand the intuition behind the above definition. 
The numerator in the ratio is the increase of $f_{i,\omg}$ when $e$ is selected. 
The denominator measures the remaining distance from the current value to $1$, and helps prioritize the scenarios that are close to being covered.
The algorithm then selects an \elem\ $e$ with the highest $G_E(e)$. 

\subsection{Estimating the Greedy Score}\label{subsec:est_greedy_score}
Since the summation in eqn.\eqref{eqn:greedy_rule} has \expo ly many terms, it is not clear how to compute the exact value of $G_E(e)$ in polynomial time. 
However, since $G_E(e)$ is the expectation of $\Delta_E(i,\omg;e)$ over the expanded scenarios, we can estimate it and select an {\em approximately} greediest \elem\ by sampling. 
The performance of this approach is guaranteed by the following result, which follows directly from the analysis in \cite{azar2011ranking} and \cite{im2016minimum}. 
\begin{theorem}[Approximate Greedy \Alg] \label{thm:sfr-apx} Let $\sigma=(e_1,\dots,e_n)$ be a \perm\ of \elem s and denote $E^t := (e_1,\dots,e_t)$ for $t\geq 1$ and $E^0:=\emptyset$.
\Sps\ for each $t=0,\dots, n-1$,  we have 
\[G_{E^t}(e_{t+1})\ge \Omega(1) \cdot \max_{e'\in [n]\bs E^t} G_{E^t}(e').\] 
Then, \[{\rm Cost}(\sigma)\le O\lb(\log \frac1\eps\rb) \cdot {\rm OPT}\] where $\rm OPT$ denotes the optimum of the SFRN problem.
\end{theorem}

\begin{algorithm}
\begin{algorithmic}[1]
\State Initialize $E \leftarrow \emptyset$ and \perm\ $\sigma=\emptyset$.
\For {$t=1,\dots,n$}
\State $E\lar \{\sigma(1),\dots,\sigma(t-1)\}$
\State For each $e\in [n]\bs E$, let $\overline{G_E}(e)$ be the \emp\ mean of $\Delta_E(i,\omg;e)$ over $N=m^3 n^4 \eps^{-1}$ independent draws of expanded \sce s $(i,\omg)$ from the \distr\ $(\pi_{i,\omg})$.
\State Let $\sigma(t)$ be the \elem\ $e\in [n]\setminus E$ that maximizes $\overline{G_E}(e)$. 
% \If{$\overline{G_E}(e) \ge \frac14 m^{-2}n^{-4}\eps$}\label{step:threshold}\Comment{If the estimated score is high}
%  \State Update $E\gets E \cup \{e^*\}$ and append $e^*$ to sequence $\sigma$.
% \Else
%  \State Exit the while loop. \qquad\qquad\qquad\qquad\qquad \Comment{Phase 1 ends}
%\EndIf
\EndFor
\State Return the \perm\  $\sigma$.
\caption{Non-adaptive SFRN algorithm\label{alg:non-adp}}
\end{algorithmic}
\end{algorithm}

To find such an approximately greediest \elem, for a fixed \elem\ $e$, we \indep ly sample a polynomial number of expanded scenarios from the \distr\ $(\pi_{i,\omg})$. 
We evaluate $\Delta_E(i,\omg,e)$ for each expanded \sce\ $(i,\omg)$ sampled, and compute their empirical mean.
Due to standard concentration bounds, the deviation from $G_E(e)$, which is its expectation, is likely small. 
Therefore, the empirical mean can serve as a reliable estimate of the greedy score.
We formally define this algorithm in Algorithm \ref{alg:non-adp}.

\subsection{Handling Small Greedy Score}\label{subsec:small_greedy_score}
The desired $O(\log \frac 1\eps)$-approximation would immediately follow if we could show that the estimation is {\em always} within a (multiplicative) $O(1)$ factor to the true score $G_E(e)$ for every \elem\ $e$. 
Unfortunately, this is {\bf not} true. 
In fact, it may fail when $G_E(e)$ is tiny for every \elem\ $e$.

To see this, consider an i.i.d. sample $X_1,\dots,X_k$ (which corresponds to $\Delta_E(i,\omg;E)$), each with mean $\mu>0$ (which corresponds to $G_E(e)$). 
Chernoff's \ineq\ states that the \prb\ of having a large deviation decays \expo ly in $k\mu$. 
In other words, to ensure a target level of confidence, we need the sample complexity $k$ to scale as $\Omg(1/\mu)$, which can be large when $\mu$ is small. 
% In other words, $\mu$ is small, we can not guarantee the quality of estimation using a 
% it is not clear whether the sampling-based approach still returns an approximately greediest element with a $k$.

To overcome this, we observe that if the score is small for {\em all} \elem s, then the set of \elem s selected so far is likely to have already covered all scenarios. 
Therefore, the choice of the next element is barely relevant.
More precisely, we show that if $\overline{G_E}(e)$ is less than a certain (small) threshold that scales polynomially in $n,m$ and $1/\eps$, then with \prb\ $1-n^{-\Omg(1)}$, the current set already covers all scenarios.
We formalize this in Lemma \ref{lem:phase2-non-adp} in Appendix \ref{apdx:nonadap}.

So far, we have explained why our algorithm (a) is efficient (in Section \ref{subsec:greedy_score}), (b) identifies a sufficiently greedy \elem\ until all scenarios are covered (in Section \ref{subsec:small_greedy_score}), and (c) leads to a low approximation factor (in Section \ref{subsec:est_greedy_score}). 
Combining the above components, we have the following main result of this section.
\begin{theorem}[Approx. Algo. for SFRN]\label{thm:non-adp-sfrn}
Algorithm \ref{alg:non-adp} is a $poly(\frac 1\eps,n,m)$ time $O(\log \frac 1\eps)$-approximation for the SFRN problem.
\end{theorem}
It should be noted that the approximation factor is best possible due to Theorem 3.1 in \citealt{azar2011ranking}.
Furthermore, observe that for the ODTN problem, we have $\eps= \frac 1 {m-1}$, so we obtain the following.
 
\begin{corollary}[Approx. Algo. for Non-\adap\ ODTN]\label{coro:non-adap-odtn}
Algorithm \ref{alg:non-adp} gives an $O(\log m)$-approximation for the non-adaptive ODTN problem where $m$ is the number of \hypos.
\end{corollary}

We defer all details to Appendix \ref{apdx:nonadap}.

\section{The ASRN Problem with Low Noise Level}\label{sec:few_star}
\def\A{\ensuremath{{\cal A}}\xspace}
\def\Ab{\ensuremath{\bar{{\cal A}}}\xspace}
\def\adaptr{{\sf ADAPT-r}\xspace}
\def\adapth{{\sf ADAPT-c}\xspace}
\def\submax{{\sf SUB-MAX}\xspace}
\def\algsubmax{{\sf ALG-SUB-MAX}\xspace}

\renewcommand{\thefootnote}{\fnsymbol{footnote}}
In this section, we present an \adap\ algorithm whose performance depends on the uncertainty level of the instance. 
Informally, \sps\ we store the response functions $\{r_i(\cdot)\}_{i\in [m]}$ as a matrix whose rows and columns correspond to the elements and scenarios. 
Then, the {\em column/row uncertainty} is the maximum number of $\star$'s in any column/row, formally defined as follows. 

\bdefn[Column and Row Uncertainty]
Given an ASRN \ins, the {\bf column uncertainty} is $c :=\max_{i\in [m]} |\{e\in [n]: r_i(e) = \star\}|$.
Similarly, the {\bf row uncertainty} is $r := \max_{e\in [n]} \{i\in [m]: r_i(e) = \star\}$. 
\edefn

The main result of this section is an $O\left(\log \frac m\eps + \min\{c\log |\Omg|,r\}\right)$-approximation for the ASRN problem for \ins s that have column uncertainty $c$, row uncertainty $r$ and separability $\eps$.
This is achieved by choosing between two \alg s, each having an approximation ratio of $O(c\log |\Omg| + \log \frac m\eps)$ and an $O(r+ \log \frac m\eps)$. 
In both algorithms, we maintain the posterior probability of each scenario based on the responses of the selected elements.
We use these probabilities to calculate a {\it score} for each element, which depends on (a) the balancedness of the partition on the remaining scenarios, resulting from selecting this element, and (b) the expected number of scenarios eliminated. 

Unlike the noiseless setting, in the ASRN (and ODTN) problem, each scenario can follow an exponential number of paths in the \dec\ tree. Therefore, a naive generalization of the analysis in \cite{navidi2016adaptive} incurs an undesirable approximation factor.

We overcome this challenge by reducing to the ASR \ins\ ${\cal J}$ defined in Section \ref{sec:exp_hypo}.
However, since ${\cal J}$ involves exponentially many scenarios, a naive implementation of the algorithm in \cite{navidi2016adaptive} leads to an exponential running time.
In Section~\ref{subsec:odtn-h} we exploit the special structure of ${\cal J}$ and devise a polynomial-time \alg.
Then, in Section~\ref{subsec:odtn-r}, we propose a slightly different algorithm than that of \cite{navidi2016adaptive}, and show an $O(r + \log \frac m\eps)$ approximation ratio.

\begin{algorithm}\label{alg:asr-odt-h}\begin{algorithmic}[1]
\State Initialize $E \leftarrow \emptyset, H' \leftarrow H$.
\While {$H'\neq \emptyset$} \State For any element $e\in [n]$, let $B_e(H')$ be the largest {\it cardinality} set among \label{algline:Be}
\begin{equation*}
\{(\ex)\in H': r_\ex(e) = o\}\, \qquad \forall o \in \Omega
\end{equation*}
\State Define $L_e(H') = H'\setminus B_e(H')$
\State Select the element $e\in [n]\setminus E$ maximizing
\begin{equation} \label{eq:asr-score}
\mathrm{Score}_c(e,E,H') = \pi\big({L}_e(H')\big) \,\,+\,\, \sum\limits_{(\ex)\in H', f_{\ex}(E)<1} \pi_{\ex} \cdot \frac{f_{\ex}(e\cup E)-f_{\ex}(E)}{1-f_{\ex}(E)}
\end{equation}
\State Observe response $o$ and update $H'$ as $H'\lar \{(\ex)\in H': \omg_e =o \mbox{ and } f_{\ex}(E\cup e)<1\}$
\State $E\leftarrow E\cup \{e\}$
\EndWhile
\caption{Algorithm for ASR instance \J, based on \cite{navidi2016adaptive} \label{alg:asr}}
\end{algorithmic}
\end{algorithm}
\subsection{An $O(c\log |\Omg|+\log \frac m\eps)$-Approximation Algorithm} 
\label{subsec:odtn-h}

Our first adaptive algorithm is based on the $O(\log \frac m\eps)$-approximation algorithm for the (noiseless) ASR problem from \cite{navidi2016adaptive}, rephrased in our notation Algorithm \ref{alg:asr}. {\new Here we emphasize that $H$ is the set of {\em expanded} \sce s; see Definition \ref{def:expanded_sce}.}
Applying this result to the ASR instance \J, we obtain an $O(\log \frac {|H|}\eps)$-\apxn.
Note that $|H| \leq |\Omg|^c\cdot m$, we immediately obtain the desired guarantee on the cost.

This algorithm maintains the set $H'\sse H$ of expanded scenarios that are compatible with all the observed outcomes, and iteratively selects the element with maximum score, as defined in~\eqref{eq:asr-score}\footnote[3]{We use the subscript $c$ to distinguish from the score function\ $\mathrm{Score}_r$
considered in Section~\ref{subsec:odtn-r}, but for ease of notation, we will suppress the subscript in this subsection.}. 
This score strikes a balance between {\it covering} the \submod\ functions (of the remaining scenarios) and {\it shrinking} $H'$ (thereby reducing the uncertainty in the target \sce).

To interpret the first term in $\mathrm{Score}_c$, for simplicity, assume that $\Omg = \{\pm 1\}$.
Upon selecting an \elem, $H'$ is split into two subsets, among which $L_e(H')$ is the lighter (in cardinality). 
% Equivalently, since $\pi_\ex$ is \unif, in the total prior probabilities.
Thus, this term is simply the number of expanded scenarios eliminated in the {\em worst} case (over the responses in $\Omg$). 
% This is reminiscent of the greedy algorithm for the ODT problem (e.g., \cite{kosaraju1999optimal}) which iteratively selects a test that maximizes the number of scenarios ruled out, in the worst case over all the test outcomes.
The higher this term, the more progress is made towards identifying the target (expanded) \sce.
The second term is similar to the score in our non-adaptive algorithm (\Alg~\ref{alg:non-adp}). 
It involves the sum of incremental coverage over all expanded scenarios, weighted by their current coverage, with higher weights on expanded scenarios closer to being covered.

As noted above, computing the summation in $\mathrm{Score}_c$ naively requires exponential time.
However, in Appendix \ref{apdx:computation_score} we explain how to utilize the structure of the ASRN instance \J to reformulate each of the two terms in $\mathrm{Score}_c$ in a manageable form, enabling a polynomial-time implementation.
Now we are ready to formally state the main result of this subsection. 
% Observe that by Theorem \ref{thm:asr}, on the ASR instance \J, Algorithm \ref{alg:asr} has an $O(\log (|\Omg|^c m) + \log \frac m \eps) = O(c\log |\Omg| + \log \frac m\eps)$ approximation ratio since $|H|\le m |\Omg|^c$.  Formally, we have the following.
\begin{theorem}[Approx. Algo., Low Column Uncertainty]\label{thm:asrn-c}
\Alg~\ref{alg:asr} can be implemented in \dxs\ time and is an $O(c\log |\Omg| +\log \frac m\eps)$-approximation algorithm for the ASRN problem on any instance with column uncertainty $c$. \label{thm:odtn_c}\end{theorem}
 
\subsection{An $O(r+\log\frac m\eps)$-Approximation Algorithm} \label{subsec:odtn-r}

%%%%%%%%%%%%%%%%%% dropped old algo %%%%%%%%%%%%%%%%%%%%
\ignore{
\begin{algorithm}\begin{algorithmic}[1]
\State initially $E \leftarrow \emptyset, H' \leftarrow H$
\While {$H'\neq \emptyset$} 
\State  $S\gets \{i\in [m]\,:\, H_i\cap H' \ne \emptyset\}$\Comment{Surviving original scenarios }
\State for each $e\in [n]$, let $\ell(e; S)$ be (one of) the lightest cardinality set among
$\{i\in S: r_i(e) = o\}, \quad \forall o\in \Omg$
\State for each $e\in [n]$, define
$R_e(H') = \{(\ex)\in H': i \in \ell(e; S)\} = H'\cap \left(\cup_{i \in \ell(e; S)} H_i \right)$
\State select test $e\in [n]\setminus E$ that maximizes
\begin{equation} \label{eq:asr-r-score}
\mathrm{Score}_r(e,E,H') = \pi\big(R_e(H')\big) \,\,+\,\, \sum\limits_{(\ex)\in H', f_{\ex}(E)<1} \pi_{\ex} \cdot \frac{f_{\ex}(e\cup E)-f_{\ex}(E)}{1-f_{\ex}(E)}
\end{equation}
\State observe outcome $o$
\State $H'\lar \{(\ex)\in H': r_\ex(e)=o\}$
\Comment{Update the inconsistent scenarios }  
% scenarios  from $H'$ based on the feedback from $e$
\State $E\leftarrow E\cup \{e\}$
\EndWhile
\caption{Modified algorithm for ASR instance \J. \label{alg:asr_*}}
\end{algorithmic}
\end{algorithm}
}
%%%%%%%%%%%%%%%%%%%%%%%%%%%%%%%%%%%%%%%%%%

In this section, we consider a slightly different score function, $\mathrm{Score}_r$, and
obtain an $O(r+\log \frac m\eps)$-approximation.
% Unlike the previous section where the approximation factor follows as an immediately corollary from Theorem~\ref{thm:asr}, to prove this result, we need to also  modify the analysis.
% The only difference lies in the first term of the score function.
Recall that in \Alg~\ref{alg:asr}, upon selecting an element $e$, the remaining expanded scenarios are partitioned into at most $|\Omg|$ subsets, where the one with the lightest cardinality is denoted $L_e(H')$.

In the {\em modified} score function $\mathrm{Score}_r$, we instead consider the partition on the {\it original} scenarios, rather than the expanded scenarios.
We define the subset $S$ of the remaining original scenarios that has at least one expanded scenario remaining.
If an element $e$ is selected, then $S$ is partitioned into (at most) $|\Omg|$ subsets, where the subset with the largest cardinality is denoted as $C_e(S)\sse [m]$.
The set $R_e(H')\sse H'$ is then defined as the consistent expanded scenarios that have a {\em different} response than $C_e(S)$.
We formally describe this score function in Algorithm \ref{alg:asr_*} in Appendix~\ref{apdx:r}. 
% we now use $q(L'_e(H'))$ instead of $q(L_e(H'))$. Therefore, as shown in \S\ref{subsec:odtn-h}, the second term in \eqref{eq:asr-r-score} (which is also the second term in \eqref{eq:asr-h-score}) equals \eqref{eq:asr-h-compact}.

Note that $S$ can be {\new maintained} efficiently. 
More generally, for each \sce\ $i$, we can efficiently maintain the number $n_i$ of expanded \sce\ of $i$ that is not eliminated.  
In fact, observe that if we select a $\star$-\elem\ $e$ for $i$, then $n_i$ decreases by half. 
Moreover, the response is incompatible with the outcome, i.e., $r_i(e)\neq o$, then $n_i$ becomes $0$.

Similarly to \Alg~\ref{alg:asr}, the main computational challenge lies in evaluating the second term, since it involves summing over exponentially many terms, but a polynomial-time implementation follows by a similar approach as outlined in Section \ref{subsec:odtn-h}.

% \begin{algorithm}
% \begin{algorithmic}[1]
% \State initially $E \leftarrow \emptyset$, $S \leftarrow [m]$ and  $p_i \leftarrow \op_i$ for all $i \in [m]$.
% \While {$|S|> 1$}
% \State for any element $e\in [n]$, let $L_e(H)$ be the smaller cardinality set among $T^+(e)\cap H$ and $T^-(e)\cap H$

% \State select test $e\in U\setminus E$ that maximizes
% {\small\begin{equation} \label{eq:alg_odt_*}
% \mathrm{Score}(e)= \mathrm{Score}_r(e,E,H') = \pi\big({L}_e(H')\big) \,\,+\,\, \sum\limits_{(\ex)\in H', f_{\ex}(E)<1} \pi_{\ex} \cdot \frac{f_{\ex}(e\cup E)-f_{\ex}(E)}{1-f_{\ex}(E),}
% \end{equation}}
% \If{outcome of test $e$ is $+$}
%  \State update $H\gets H\setminus T^-(e)$
% \Else
%  \State update $H\gets H\setminus T^+(e)$
% \EndIf

% \State $E\leftarrow E\cup \{e\}$
% \For{$x\in H$}
% \If{$r_x(e) = *$}
%     \State $p_x\leftarrow p_x/2$ \EndIf\EndFor
% \EndWhile
%  \caption{Polynomial time implementation. \label{alg:odt_*}}
%  \end{algorithmic}
% \end{algorithm}

The main result of this section, stated below, is proved by adapting the proof technique from \cite{navidi2016adaptive} and formally proved in Appendix~\ref{apdx:r}.

\begin{theorem}[Apxn. Algo., Low  Row Uncertainty]\label{thm:odtn-r}
Algorithm \ref{alg:asr_*} can be implemented in \dxs\ time and is an $O(r+\log \frac m\eps)$-approximation algorithm for the ASRN problem for any instance with row uncertainty $r$.
\end{theorem}

By selecting between \Alg~\ref{alg:asr} and \Alg~\ref{alg:asr_*} depending on whether $c\log |\Omg| > r$, we immediately obtain the following.
\begin{theorem}[Meta Algo. for ASRN]\label{thm:asrn}
There is an adaptive $O(\min\{c\log |\Omg|,r\}+ \log \frac m\eps)$-approximation algorithm for the ASRN problem.
\end{theorem}
In \parti, this gives an $O(\min\{c\log |\Omg|, r\}+ \log \frac m\eps)$-approximation algorithm for the ODTN problem. 
We also provide closed-form expressions for the scores used in Algorithm \ref{alg:asr} and Algorithm \ref{alg:asr_*} for the ODTN problem in Appendix~\ref{app:odtn-score}, which will be used for our experiments.

\section{ODTN with Many Unknowns}\label{sec:many_star}
% \su{check whether I missed any lemma statement, and rename the apdx section "missing pfs in sec xx"}
Our adaptive algorithm in Section \ref{sec:few_star} has a low \apxn\ ratio when the vast majority of entries in the test-\hypo\ \mtx\ are deterministic.
In this section, we focus on the other extreme, where ODTN instance has very {\em few} deterministic outcomes.

More precisely, we quantify the noise level by its {\em sparsity}. 
An ODTN \ins\ is $\alpha$-sparse if every test has $O(m^\alpha)$ deterministic \hypos.
Our main result is a polynomial-time \apxn\ algorithm with cost  $O(m^\alpha + \log m \cdot {\rm OPT})$ where ${\rm OPT}$ is the optimum of the ODTN problem, whose output may be wrong with a low \prb.
Furthermore, we show that for any $\alpha \in [0,1]$ we have ${\rm OPT} = \Omg(m^{1-\alpha})$. 
Therefore, when $\alpha < \frac 12$, we obtain an $O(\log m)$-\apxn\ for the ODTN problem. 
It should be noted that the cost matches the APX-hardness result (Theorem 4.1 of \citealt{ChakaravarthyPRAM11}) within $O(1)$ factors. 
We next explain the ideas in more detail.

\subsection{Stochastic Set Cover}
% As our analysis involves intricate technicalities, we provide a broad overview in this subsection. 
% Sections \ref{subsec:main} and
% \ref{subsec:member-overview} are dedicated to elaborating on the individual steps outlined here and can be skipped during the first reading.
The design and analysis of our algorithm are closely related to the problem of {\it Stochastic Set Cover} (SSC) (\citealt{LiuPRY08,im2016minimum}).
An SSC instance consists of a {\it ground set} $[m]$ of {\it items}
and a collection of {\it random} subsets $S_1,\cdots, S_n$ of $[m]$. 
The distribution of each subset is known, but its instantiation is unknown until being selected.
The goal is to minimize the expected number of sets to cover all \elem s in $[m]$.

A key component of our analysis is the following lower bound on the optimum of the ODTN problem, in terms of the optima of the following SSC instances.
Recall that a test $T$ can be represented as a three-way partition $(T^+, T^-, T^*)$ of $[m]$.

\bdefn[Induced SSC Instances]{}
For any \hypo\ $i\in [m]$, let $\mathrm{SSC}(i)$ denote the SSC instance with ground set $[m]\setminus \{i\}$ and $n$ random sets, given by
\[S_T(i) = \left\{ \begin{array}{ll}
T^+\mbox{ with prob. }1 & \mbox{ if }i\in T^-  \\
T^-\mbox{ with prob. }1 & \mbox{ if }i\in T^+  \\
T^- \mbox{ or } T^+ \mbox{ with prob. $\frac12$ each} & \mbox{ if }i\in T^*  
\end{array}\right.,\qquad \forall T\in [n].\]
\edefn

To see the connection between the SSC and ODTN problem, observe that when $i$ is the target \hypo\ in the ODTN \ins, any feasible algorithm must identify $i$ by {\it eliminating} all other \hypos. 
In the SSC terminology, we have {\it covered} all items in $[m]\bs \{i\}$.
This leads to the following lower bound.
\begin{proposition}[SSC-based Lower Bound]\label{lem:lower-bound} 
For any ODTN \ins\ with optimum OPT and induced SSC \ins es $\{{\rm SSC}(i)\}_{i\in [m]}$, we have 
\[\mathrm{OPT}\geq \sum_{i\in [m]} \pi_i \cdot \mathrm{OPT}_{\mathrm{SSC}(i)}.\]
\end{proposition}

Therefore, to bound the cost of an ODTN \alg, we only need to charge its cost to the corresponding SSC instances and apply the above \ineq.
The next two subsections are dedicated to constructing such an \alg. 

\subsection{A Greedy SSC \Alg}
A natural greedy algorithm is known to be an $O(\log m)$-approximation (\citealt{LiuPRY08,im2016minimum}). 
As we recall, the greedy algorithm for the (deterministic) Set Cover problem iteratively selects a set that covers the largest number of new items (i.e., items that are not covered so far).
Analogously, in the SSC problem, the greedy \alg\ selects the set that maximizes the {\em expected} number of new items covered. 

We will consider an even more general version of the greedy algorithm, dubbed {\em $(\beta,\rho)$-greedy} where $\beta,\rho>1$ are constants. 
{\new This \alg\ applies the greedy rule for an $\Omg(1/\rho)$ fraction of iterations. 
Furthermore, in those iterations, instead of implementing the exact greedy rule, it selects a set whose coverage is $\Omg(1/\beta)$ fraction that of the greediest set. 
To formalize, for any deterministic subset $U\sse [n]$ and a random subset $S$, we denote by ${\rm Cov} (S;U) = \ho{E}_S [|S\bs U|]$ the {\em expected  coverage}.

\bdefn[$(\beta,\rho)$-greedy] An \alg\ is {\bf $(\beta,\rho)$-greedy} for the SSC problem, if the (random) \xulie\ of sets $S_1,S_2,\dots$ it selects \sat\ the following with \prb\ $1$: For any $t\ge 1$, there is a subset $I\sse \{1,\dots, t\}$ with $|I|\ge t/\rho$, such that for any $i\in I$, we have \[{\rm Cov}(S_i; S_1\cup\dots \cup S_{i-1}) \ge \frac 1\beta \max_{S\in {\cal C}} {\rm Cov}(S; S_1\cup\dots \cup S_{i-1}).\]
\edefn }

\iffalse 
\begin{algorithm}
\begin{algorithmic}[1]
\State Initialize $\mathcal{C} \lar \emptyset, U \leftarrow [m]$.\Comment{Selected sets and uncovered items}
\While {$U\ne \emptyset$} 
\For{$i\notin \mathcal{C}$}
\State ${\rm Cov}(i;U)\lar \ho{E}[U\cap S_i]$
\Comment{Compute the expected coverage of $S_i$}
\EndFor
\State Select any $i^*\notin \mathcal{C}$ \sat ing ${\rm Cov}(i^*;U) \ge \frac 1\beta \max_{i\in \mathcal{C}} {\rm Cov}(i;U)$
\State Observe the instantiation $\bar S_{i^*}$ of $S_{i^*}$
\State $U\lar U\bs \bar S_i$ \Comment{Update the uncovered items}
\State $\mathcal{C} \lar \mathcal{C}\cup \{i^*\}$
\EndWhile
\caption{$(\beta,\rho)$-Greedy algorithm for the SSC Problem}
\label{alg:ssc_greedy}
\end{algorithmic}
\end{algorithm} 
\fi

The following is implied by Theorem 1.1 of \citealt{im2016minimum} and serves as the cornerstone for our analysis.
\begin{theorem}[Greedy SSC Algorithm]\label{thm:ssc} 
Any $(\beta,\rho)$-greedy algorithm with $\beta,\rho>1$ is an $O(\beta \rho \log m)$-\apxn\ for the SSC problem.
\end{theorem}

This result inspires a simple greedy \alg\ for the ODTN problem, which we describe in the next subsection and use as a strawman to motivate further algorithmic ideas.

\subsection{A First Attempt: SSC-based Greedy ODTN \Alg}
Our ODTN algorithm is inspired by the following key observation. 
\Sps\ $A$ is the set of alive (i.e., not yet eliminated) \hypos\ in the ODTN problem, and a test $T$ maximizes $|T^+ \cap A| + |T^- \cap A|$. 
Then, $T$ results in good progress for all SSC instances ${\rm SSC}(i)$ with $i\in T^*$ {\em simultaneously}.

\begin{lemma}[Greedy Is Good for Most Hypotheses]\label{prop-greedy-star}
Let $A\sse [m]$ and $T$ be a test such that  
\begin{align}\label{eqn:110523}
\E\left[|S_T(i)\cap (A\bs i)|\right] = \frac 12 \left( |T^+\cap A| + |T^-\cap A|\right)\ge \max_{T'\in [n]} \frac 12 \left(|(T^+)'\cap A| + |(T^-)' \cap A|\right).
\end{align}
Then, for any \hypo\ $i\in T^*$, we have
$$ \E\left[|S_T(i)\cap (A\bs i)|\right] \, \ge\, \frac12 \cdot \max_{T'\in [n]} \E\left[|S_{T'}(i)\cap (A\bs i)|\right].$$
\end{lemma}

% To see this, fix an \arb\ \hypo\ $i$, and consider a test $T_i$ that maximizes the expected coverage in the SSC instance $\mathrm{SSC}(i)$. 
% Formally, we have the following \ineq, whose proof is \strfwd.
% \Sps\ $A$ is  the set of remaining \hypos.
% Then, the expected coverage of the random set $S_T(i)$ is $\frac 12\left( |T^+\cap A| + |T^-\cap A|\right)$. 
% Therefore, a test $T$ that maximizes $\frac 12( |T^+\cap A| + |T^-\cap A|)$ is $2$-greedy for $\mathrm{SSC}(i)$.

It should be noted that, in general, the above does not hold for $i\notin T^*$. 
To see this, \sps\ a test $T$ \sats\ eqn. \eqref{eqn:110523} and has imbalanced deterministic sides, for example, $|T^+|=m^\alpha$ and $|T^-| = 1$. 
Then, for each $i\in T^+$, the random set $S_T$ has poor coverage in the SSC instance ${\rm SSC}(i)$, since it covers only one item (w.p. $1$). 

By Lemma \ref{prop-greedy-star}, a test $T$ makes good progress for {\em most} SSC instances if (i) $T$ \sats\ eqn. \eqref{eqn:110523}, and (ii) $i\in T^*$ is \satd\ for most \hypos\ $i$. 
This motivates us to consider the class of ODTN instances where (ii) is satisfied.

\bdefn[Sparse Instance] An ODTN instance is {\bf $\alpha$-sparse} for some $\alpha \in [0,1]$ if for all tests $T\in \T$ we have $\max\{|T^+|,|T^-|\}\leq  m^{\alpha}$.
\edefn

By this definition, if an ODTN \ins\ is  $\alpha$-sparse, then most \hypos\ are in $T^*$. 
Consequently, by Lemma \ref{prop-greedy-star}, a test $T$ \sat ing eqn. \eqref{eqn:110523} is $2$-greedy for most (more precisely, $m-O(m^\alpha)$) SSC \ins s. 

This motivates the following naive greedy algorithm. 
\Sps\ $A$ is the set of consistent \hypos. 
In each iteration, we select a test $T$ that maximizes $\frac 12 |T^+\cap A| + \frac12 |T^-\cap A|$. 
Furthermore, consider the 
following {\em ideal event}:
\begin{center}
For every $t\ge 1$, when we select the $t$-th test, for {\em every} hypothesis $i\in [m]$, the algorithm has selected $\Omg(t/\rho)$ tests $T$ with $i\in T^*$. 
\end{center}
If this event occurs, then the naive greedy \alg\ gives an $O(\log m)$-\apxn.
In fact, since the \xulie\ of tests selected is $(2,\rho)$-greedy for every $i$, by Theorem~\ref{thm:ssc}, the expected cost conditional on $i$ being the true \hypo\ is $O(\rho\log m)\cdot \mathrm{OPT}_{\mathrm{SSC}}(i)$.
Taking the expectation over all \hypos\ and combining with the SSC-based lower bound in Proposition ~\ref{lem:lower-bound}, we deduce that the total cost is $O(\rho \cdot \log m) {\rm OPT}$.

However, the ideal event may not always occur. 
Next, we explain how to fix this problem by intermittently enumerating a small subset of \hypos\ with the highest posterior probabilities.

\subsection{Last Piece of the Puzzle: the Membership Oracle}
\Sps\ the ideal event fails, that is, up until some iteration, the \xulie\ of tests selected is no longer $(2,\rho)$-greedy for some \hypo\ $i$.
To handle this issue, we modify the above greedy algorithm as follows: 
For each iteration $t=2^k$ where $k=1,2,\dots,\log m$, we consider the set $Z = Z_k$ of $O(m^\alpha)$ hypotheses with the fewest $\star$-tests selected so far. 
Equivalently, we may maintain a posterior \prb\ using Bayes' rule and define $Z$ as the subset of $O(m^\alpha)$ \hypos\ with the highest posterior probabilities.

Then, we invoke a {\it membership oracle} ${\rm Member}(Z)$ to check whether the target \hypo\ $\bar{i}\in Z$.
If so, then the algorithm terminates and returns $\bar{i}$. 
Otherwise, it continues with the greedy algorithm until the next power-of-two \iter.

Specifically, the membership oracle Member$(Z)$ takes a subset $Z\sse [m]$ of \hypos\ as input, and decides whether the target \hypo\ $\bar{i}$ is in $Z$. 
% At a high level, $\mathrm{Member}(Z)$ works as follows. 
Whenever $|Z|\geq 2$, we pick an \arb\ pair $(j,k)$ of \hypos\ in $Z$ and choose a test where these two \hypos\ have distinct \dtmnstc\ outcomes. 
Such a test exists due to Assumption \ref{assu:iden}.
Moreover, since each of these tests rules out at least one \hypo, within  $|Z|-1$ \iter s, there is only one \hypo\ left.
In Appendix \ref{subsec:member}, we explain how to verify whether this remaining \sce\ is the true \hypo\ using $O(\log m)$ tests.

We can bound the cost of the membership oracle as follows.
\begin{lemma}[Membership Oracle Has Linear Cost]
\label{prop:membership_oracle}
If $\bar{i} \in Z$, then $\mathrm{Member}(Z)$ 
declares $\bar{i} = i$ with probability $1$; otherwise, it 
declares $\bar{i}\notin Z$ with probability $1- O(m^{-2})$. Furthermore, the expected cost of $\mathrm{Member}(Z)$ 
 is $O(|Z|+ \log m)$.
\end{lemma}
The formal proof of the above result is deferred to Appendix \ref{subsec:member}.
At this juncture, we have introduced all relevant concepts and ideas. In the next subsection, we formally state our results. 

\subsection{Sparsity-dependent Approximation \Alg} 
We are now ready to state the overall algorithm in \Alg~\ref{alg:large-noise}. 
In each iteration, we maintain a subset of consistent \hypos, and iteratively compute the greediest test, as formally specified in Step \ref{step:alg-many-star-2}.
At each power-of-two iteration $t=2^k$ where $k=1,2,\dots,\log m$, we invoke the membership oracle and terminate if it declares a true \hypo.
This \alg\ has the following guarantee. 

\begin{algorithm}
\begin{algorithmic}[1]
\State Initialization: consistent hypotheses $A\lar [m]$, weights $w_i\lar 0$ for $i\in [m]$, \iter\ index $t\leftarrow 0$ 
\While{$|A|>1$}
\If{$t$ is a power of 2}
\State \label{step:alg-many-star-1}Let $Z\sse A$ be the subset of $2m^\alpha$ hypotheses with lowest $w_i$
\State Invoke Member$(Z)$
\State If a hypothesis is identified in $Z$, then Break
\EndIf
\State \label{step:alg-many-star-2} Select a test $T\in \T$ maximizing 
$\frac{1}{2}(|T^+\cap A| + |T^- \cap A|)$ 
\State Observe outcome $o_T$
\State $R\lar \{i\in [m]: M_{T,i}= -o_T\}$ and $A\leftarrow A\bs R$
\Comment{Remove incompatible hypotheses}
\State $w_i \leftarrow w_i+1$ for each for each $i\in T^*$ \Comment{Update the weights of the \hypos}
\State $t\leftarrow t+1$.
\EndWhile
\caption{Main algorithm for large number of noisy outcomes\label{alg:large-noise}}
\end{algorithmic}
\end{algorithm}

\begin{theorem}[Apxn. Algo. for Sparse Instances]\label{thm:sparse} \Alg\ \ref{alg:large-noise} is a polynomial-time algorithm which (a) has cost $O(m^\alpha + \log m\cdot {\rm OPT})$ for any $\alpha$-sparse instance with $\alpha \in [0,1]$, where $\rm OPT$ is the optimum for the ODTN problem, and (b) returns the true \hypo\ with \prb\ $1-m^{-1}$.
\end{theorem}

The choice of $m^{-1}$ is not essential: To reduce the error \prb, we can simply repeat the algorithm many times and perform a majority vote, i.e., return the most frequent output.
Next, we argue that the first term, $m^\alpha$, is negligible compared to OPT when $\alpha\le \frac 12$.

\begin{proposition}[Sparsity-based Lower Bound on OPT]
\label{lem:sparse-OPT}
For any $\alpha$-sparse instance, we have $\mathrm{OPT}= \Omega(m^{1-\alpha})$.
\end{proposition}
In \parti, when $\alpha<\frac 12$ the above implies that the cost $O(m^\alpha)$ for each call of the membership oracle is lower than OPT, and therefore the total cost incurred in the power-of-two steps is $O(\log m \cdot \mathrm{OPT})$. 
We therefore conclude the following.

\bcoro[Logarithmic-Approximation for Sparse Instances]
\Alg\ \ref{alg:large-noise} has cost $O(\log m \cdot {\rm OPT})$ for any ODTN \ins\ with $\alpha \le \frac 12$ and returns the true \hypo\ with \prb\ $1-m^{-1}$. \ecoro

\subsection{Analysis Outline}
We outline the proof of Theorem \ref{thm:sparse} and defer the formal proof to Appendix \ref{apdx:many_stars}.

\noindent{\bf Truncated Decision Tree.} 
Let $\mathbb{T}$ denote the decision tree corresponding to our algorithm. 
We only consider tests that correspond to step~\ref{step:alg-many-star-2}.
Recall that $H$ is the set of {\it expanded} hypotheses and that any expanded \hypo\ traces a unique path in $\mathbb{T}$.
For any $(\ex)\in H$, let $P_\ex$ denote this path traced;
so $|P_\ex|$ is the number of tests performed in Step \ref{step:alg-many-star-2} under $(\ex)$. 
We will work with a truncated decision tree $\overline{\mathbb{T}}$, defined below. 

Fix any expanded \hypo\ $(\ex)\in H$. 
For any $t\ge 1$, let $\theta_\ex(t)$ denote the fraction of the first $t$ tests in $P_\ex$ that are $\star$-tests for hypothesis $i$. 
Recall that $P_\ex$ only contains tests from Step~\ref{step:alg-many-star-2}. 
Let $\rho=4$ and define
\begin{equation}\label{def:trunc}
t_\ex = \max\left\{ t \in \{2^0, 2^1,\cdots, 2^{\log m}\} \,\,:\,\, \theta_\ex (t') \ge \frac 1\rho \mbox{ for all }t'\le t \right\}.
\end{equation}
If $t_\ex > |P_\ex|$ then we simply set $t_\ex = |P_\ex|$. 

Now we define the {\it truncated} decision tree $\overline{\mathbb{T}}$.
By abuse of notation, we will use $\theta_i(t)$ and $t_i$ as {\it \rv s}, with randomness over $\omg$.
Observe that for any $(\ex)$, at the next power-of-two step\footnote{Unless stated \ow, we denote $\log := \log_2$.}
$2^{\lceil\log t_i\rceil}$, which we call the {\it truncation time}, the membership oracle will be invoked. Moreover, $2^{\lceil\log t_i\rceil}\leq 2t_i$, .
This motivates us to define $\overline{\mathbb{T}}$
is the subtree of $\mathbb{T}$ consisting of the first $2^{\lceil \log t_\ex\rceil}$ tests along path $P_\ex$, for each $(\ex)\in H$.
Under this definition, the cost of \Alg~\ref{alg:large-noise} clearly equals the sum of the cost the truncated tree and cost for invoking membership oracles.

Our proof proceeds by bounding the cost of \Alg~\ref{alg:large-noise} at power-of-two steps and other steps. 
In other words, we will decompose the cost into the cost incurred by invoking the membership oracle and selecting the greedy tests. 
We start with the easier task of bounding the cost for the membership oracle.
% Note that the membership oracle is called for $O(\log m)$ times as the total number $t$ of tests used is always at most $O(m)$.
The oracle Member is always invoked on $|Z|=O(m^\alpha)$ hypotheses. 
Using Lemma~\ref{prop:membership_oracle}, the expected total number of tests due to Step~\ref{step:alg-many-star-1} is $O(m^\alpha\log m)$. 
By Lemma~\ref{lem:sparse-OPT}, when $\alpha \leq \frac 12$, this cost is $O(\log m \cdot \mathrm{OPT})$.

The remaining part of this subsection focuses on bounding the cost of the truncated tree as $O(\log m)\cdot \mathrm{OPT}$.
With this \ineq, we obtain an expected cost  of 
$$O(\log m) \cdot (m^\alpha + {\rm OPT}) \le_{(\mbox{as }\alpha<\frac12)}\, O(\log m) \cdot (m^{1-\alpha} +  {\rm OPT})  \le_{\big(\mbox{Lemma}~\ref{lem:sparse-OPT}\big)} O(\log m)\cdot {\rm OPT},$$
and Theorem~\ref{thm:sparse} follows.
At a high level, for a fixed \hypo\ $i\in [m]$, we will bound the cost of the truncated tree as follows:
\begin{align*} 
&\text{\quad\quad\quad\ \ } i \text{ has low fraction of $\star$-tests at } t_i\\
& \underset{Lemma~\ref{lem:markov}} \Longrightarrow i \text{ is among the top } O(m^\alpha) \text{ \hypos\ at } t_i\\ 
&\underset{Lemma~\ref{prop:membership_oracle}} \Longrightarrow i \text{ is identified w.h.p. by } \mathrm{Member}(Z) \text{ at } 2^{\lceil\log t_i \rceil} \leq 2t_i,\\ &\quad \quad\quad \quad \text{ hence the truncated path is } (2,2)\text{-greedy}\\
& \underset{Theorem~\ref{thm:ssc}} \Longrightarrow \text{ the expected cost \cond al on } i \text{ is } O(\log m)\cdot \mathrm{SSC}(i)
\end{align*}
and finally by summing over $i\in [m]$, it follows from Lemma~\ref{lem:lower-bound} that the cost of the {\it truncated} tree is $O(\log m)\cdot$OPT. 
We formalize each step below.

% \noindent{\bf Relating costs of $\mathbb{T}$ and  $\overline{\mathbb{T}}$.} 
% We now show that the cost of $\overline{\mathbb{T}}$ is at most twice that of $\mathbb T$.
% Intuitively, at the next power-of-two step after \iter\ $\tau$, which is within another $\tau$ \iter s, the membership oracle will identify $i$ as the target \hypo\ and algorithm halts.
% Thus, the number of iterations {\it after} \iter\ $\tau$ is at most $\tau$, and hence the expected cost of $\mathbb{T}$ is at most twice that of $\overline{\mathbb{T}}$.
% \begin{lemma}\label{lem:abar-to-a}
% For each $x\in H$, the number of tests performed $|P_x|\le 2\cdot t_x$. 
% Hence, the expected cost of $\mathbb{T}$ is at most twice that of $\overline{\mathbb{T}}$.
% \end{lemma}
Consider the first step, formally we show that if $\theta_i(t)< \frac 14$, then there are $O(m^\alpha)$ \hypos\ with fewer $\star$-tests than $i$.
% The most crucial step is showing that at any power-of-two step, there are very few \hypos\ $i$ with $\theta_i(t)\leq \frac 12$.
\Sps\ $i$ is the target \hypo\ and $\theta_i(t)$ drops below $\frac 14$ at $t$, that is, only less than a quarter of the tests selected are $2$-greedy for $\mathrm{SSC}(i)$.
Recall that if $i\in T^*$ where $T$ maximizes $\frac 12 (|A\cap T^+| +|A\cap T^-|)$, then $S_T(i)$ is $2$-greedy set for $\mathrm{SSC}(i)$, so we deduce that less than  a $\frac t4$ tests selected are $\star$-tests for $i$, or, at least $\frac {3t}4$ tests selected thus far are {\it \dtmnstc} for $i$.
We next utilize the sparsity \assu\ to show that there can be at most $O(m^\alpha)$ such \hypos.
% Therefore, to bound the number of \hypo\ $i$ with $\theta_i(t)< \frac 14$, it suffices to bound the number of \hypos\ with at least $\frac {3t}4$ tests, as formalized below.
% \footnote{Readers may find this somewhat analogous to Markov's \ineq\ in \prb\ theory.}
\begin{lemma}\label{lem:markov}
Consider any $W\sse \T$ and $I\sse [m]$. 
For $i\in I$, let $D(i) = |\{T\in W : M_{T,i}\ne *\}|$ denote the number of tests in $W$ for which $i$ has deterministic (i.e. $\pm 1$) outcomes.
For each $\kappa\geq 1$, define $I'=\{i\in I : D(i) > |W|/\kappa\}$.
Then, $|I'|\leq \kappa m^\alpha$. \end{lemma}
\begin{proof} By definition of $I'$ and $\alpha$-sparsity, it holds that 
$$|I'|\cdot \frac{|W|}\kappa < \sum_{i\in I} D(i) = \sum_{T\in W} |\{i\in I : M_{T,i}\ne *\}| \le |W|\cdot m^\alpha, \quad \quad$$
where the last step follows since $|T^*|\leq m^\alpha$ for each test $T$.
The proof follows immediately by rearranging. \end{proof}

% \noindent{\bf Bounding cost of $\overline{\mathbb{T}}$.} 
We now complete the analysis using the relation to SSC. 
Fix any hypothesis $i\in [m]$ and consider the decision tree $\overline{\mathbb{T}}_i$ obtained by {\em conditioning} $\overline{\mathbb{T}}$ on $\bar{i}=i$. 
Lemma~\ref{prop-greedy-star} and the definition of truncation together imply that $\overline{\mathbb{T}}_i$ is $\left(2,4\right)$-greedy for $\mathrm{SSC}(i)$, so by Theorem~\ref{thm:ssc}, the expected cost of $\overline{\mathbb{T}}_i$ is $O(\log m)\cdot \mathrm{OPT}_{\mathrm{SSC}(i)}$. 
Now, taking expectations over $i\in [m]$, the expected cost of $\overline{\mathbb{T}}$ is $O(\log m) \sum_{i=1}^m \pi_i\cdot \mathrm{OPT}_{\mathrm{SSC}(i)}$. 
Recall from Proposition \ref{lem:lower-bound} that \[\mathrm{OPT}\geq \sum_{i\in [m]} \pi_i \cdot \mathrm{OPT}_{\mathrm{SSC}(i)},\] 
and therefore the cost of $\overline{\ho {T}}$ is $O(\log m)\cdot \mathrm{OPT}$. 
%This bounds the cost dues to tests in Step~\ref{step:alg-many-star-2}. 
%This completes the proof of Theorem~\ref{thm:sparse}.

\noindent{\bf Correctness.} We finally show that our algorithm identifies the target \hypo\ $\bar i$ with high \prb.
By the definition of $t_i$, where the path is truncated, $\bar i$ has less than $\frac 14$ fraction of $\star$-tests. 
Thus, at \iter\ $2^{\lceil \log t_{\bar i} \rceil}$, i.e., the first time the membership oracle is invoked after $t_i$, $\bar i$ has less than $\frac 12$ fraction of $\star$-tests.
Hence, by Lemma~\ref{lem:markov}, $\bar i$ is among the $O(m^\alpha)$ \hypos\ with fewest $\star$-tests.
Finally it follows from Lemma~\ref{prop:membership_oracle} that $\bar{i}$ is identified correctly with probability at least $1-\frac1m$. 

{\new \brmk Unfortunately, Theorem \ref{thm:sparse} cannot be extended to the ASRN setting unless we impose extra \assu s on the instance. 
Essentially, our analysis crucially  relied on Lemma \ref{prop-greedy-star}, which states that the greediest set makes a good progress {\bf simultaneously} to most SSC instances, provided the ODTN \ins\ is sparse. 
However, we are not sure how to translate this property to a natural condition in the ASRN setting. 
\ermk }

\section{Extension to Non-identifiable ODT Instances}\label{app:non-id}

Previous work on ODT problem usually imposes the following {\it identifiability} \assu\ (e.g. \cite{kosaraju1999optimal}): for every pair \hypos, there is a test that distinguishes them deterministically. 
However in many real world applications, such \assu\ does not hold.
% Thus far, we have also made this identifiability assumption for ODTN (see \S\ref{subsec:odtn}). 
In this section, we explain how our results can be extended also to non-identifiable ODTN instances. 

%Our $O(\min\{r,c\log |\Omg|\} + \log m)$-\apxn\ on ODTN, implied as corollaries of Theorem~\ref{thm:asrn}, can be extended to handle this setting by 
To this end, we introduce a slightly different  stopping criterion for non-identifiable instances. (Note that is is no longer possible to stop with a unique compatible hypothesis.) 
Define a {\em similarity graph} $G$ on $m$ nodes, each corresponding to a hypothesis, with an edge $(i,j)$ if there is {\em no} test separating $i$ and $j$ deterministically. Our algorithms' performance guarantees will now also depend on the maximum degree $d$ of $G$; note that $d=0$ in the perfectly identifiable case.
For each hypothesis $i\in [m]$, let $D_i\sse [m]$ denote the set containing $i$ and all its neighbors in $G$. 
We now define two stopping criteria.
\begin{itemize}
\item {\bf Neighborhood stopping criterion:} Stop when the set $K$ of compatible hypotheses is contained in {\em some} $D_i$, where $i$ might or might not be the true hypothesis $\bar{x}$.   
\item {\bf Clique stopping criterion:}
Stop when $K$ is contained in some clique of $G$. 
\end{itemize}
Note that clique stopping is clearly a stronger notion of identification than  neighborhood stopping. 
That is, if the clique-stopping criterion is satisfied then so is  the neighborhood-stopping criterion. 
We now   obtain 
%a non-adaptive algorithm with approximation ratio $O((d+1)\log m)$ for neighborhood-stopping and 
an adaptive algorithm with approximation ratio $O(d+\min(h,r)+\log m)$ for clique-stopping as well as  neighborhood-stopping. 

Consider the following two-phase algorithm. 
In the first phase, we will identify some subset $N\sse [m]$ containing the realized hypothesis $\bar{i}$ with $|N|\le d+1$. 
Given an ODTN instance with  $m$ hypotheses and tests $\T$, we  construct  the following ASRN instance with hypotheses  as scenarios and tests as elements (this is similar to the construction in \S\ref{subsec:asr-noise}). 
The responses are the same as in ODTN: so the outcomes $\Omega=\{+1,-1\}$. Let  $U = \T\times \{+1,-1\}$ be the element-outcome pairs. For each hypothesis $i\in [m]$, we define a submodular function:
$$\widetilde{f}_i(S) = \min \left\{ \frac1{m-d-1}\cdot \big| \bigcup_{T: (T,+1)\in S} T^- \,\bigcup \,\bigcup_{T: (T,-1)\in S} T^+\big| \,,\, 1\right\},\quad \forall S\sse U.$$
It is easy to see that each function $\widetilde{f}_i: 2^U \rightarrow [0,1]$ is monotone and submodular, and  the separability parameter $\eps =\frac{1}{m-d-1}$. 
Moreover, $\widetilde{f}_i(S)=1$ if and only if at least $m-d-1$ hypotheses  are incompatible with at least one outcome in $S$. Equivalently, $\widetilde{f}_i(S)=1$ iff there are at most $d+1$ hypotheses compatible with $S$.  By definition of graph $G$ and max-degree $d$, it follows that  function $\widetilde{f}_i$ can be covered (i.e. reaches value one) irrespective of the noisy outcomes. Therefore, by Theorem~\ref{thm:asrn} we obtain an $O(\min(r,c) + \log m)$-approximation algorithm for this ASRN instance. Finally, note  that any feasible policy for ODTN with clique/neighborhood stopping is also feasible for this ASRN instance. So, the expected cost in the first phase is $O(\min(r,c) + \log m)\cdot OPT$. 

Then, in the second phase, we run a simple splitting algorithm that iteratively selects any test $T$ that splits the current set $K$ of consistent hypotheses (i.e., $T^+\cap K \ne \emptyset$ and $T^-\cap K \ne \emptyset$).  
The second phase continues until $K$ is contained in (i) some clique (for clique-stopping) or (ii) some subset $D_i$ (for neighborhood-stopping). 
Since the number of consistent hypotheses $|K|\le d+1$ at the start of the second phase, there are at most $d$ tests in this phase. So, the expected cost is at most $d \le  d \cdot OPT$.
Combining both phases, we obtain the following.
\begin{theorem}[Apxn. Algo. for Non-identifiable Instances]
There is an adaptive $O(d+\min(c,r)+\log m)$-approximation algorithm for the ODTN problem with the clique-stopping or neighborhood-stopping criterion.
\end{theorem}

\section{Experiments}
\label{sec:experiments}
We implemented our algorithms  on real-world and synthetic data sets. 
We compared our algorithms' cost (expected number of tests) with an information theoretic lower bound on the optimal cost and show that the difference is negligible. 
Thus, despite  our logarithmic approximation ratios, the practical performance is much better.

\noindent{\bf Chemicals with Unknown Test Outcomes.}
%One natural application of ODT is identifying chemical or biological materials. 
We considered a data set called WISER\footnote{https://wiser.nlm.nih.gov},  which includes 414 chemicals (hypothesis) and 78 binary tests. Every chemical has either positive, negative or unknown result on each test. The original instance (called WISER-ORG) is not identifiable: so our result does not apply directly. In Appendix~\ref{app:non-id} we show how our result can be extended to such ``non-identifiable'' ODTN instances (this  requires a more relaxed stopping criterion defined on the ``similarity graph''). In addition, we also generated a modified dataset by removing chemicals that are not identifiable from each other, to obtain a  perfectly identifiable dataset (called WISER-ID). In generating the WISER-ID instance,  we used a greedy rule that iteratively drops the highest-degree hypothesis in the similarity graph until all remaining hypotheses are uniquely identifiable. WISER-ID has 255 chemicals.

%So, we  We have performed our algorithms on both the original instance, in which some chemicals are not , and a modified version. In the modified version, to ensure every pair of chemicals can be distinguished, we removed the chemicals that are not identifiable from each other to obtain WISER-ID dataset with 255 chemicals ().

%( and we call it WISER-ORG here),

\noindent{\bf Random Binary Classifiers with Margin Error.} 
We construct a dataset containing 100 two-dimensional points, by picking each of their attributes uniformly in $[-1000, 1000]$. We also choose 2000 random triples $(a,b,c)$ to form linear classifiers $\frac{ax+by}{\sqrt{a^2+b^2}}+c\le 0$, where  $a,b\sim N(0,1)$ and $c\sim U(-1000, 1000)$. The point labels are binary and we introduce noisy outcomes based on the distance of each point to a classifier. Specifically, for each threshold $d\in \{0, 5, 10, 20, 30\}$ we define dataset CL-$d$ that has a noisy  outcome for any classifier-point pair where the distance of the point to the boundary of the classifier is smaller than $d$. 
In order to ensure that the instances are perfectly identifiable, we remove  ``equivalent'' classifiers and we are left with 234 classifiers.

\noindent{\bf Distributions.} 
For the distribution over the hypotheses, we considered permutations of power law distribution ($\Pr[X=x;\alpha]= \beta x^{-\alpha}$) for $\alpha = 0, 0.5$ and 1. 
Note that, $\alpha = 0$ corresponds to uniform distribution. 
To be able to compare the results across different classifiers' datasets meaningfully, we considered the same permutation in each distribution.

\noindent{\bf Algorithms.} We implement the following algorithms: the adaptive $O(r+\log m + \log \frac 1\eps)$-approximation (which we denote $\textrm{ODTN}_r$), the adaptive $O(c\log |\Omg| \log \frac m\eps)$-approximation  ($\textrm{ODTN}_c$), the non-adaptive $O(\log m)$-approximation (Non-Adap) and a slightly adaptive version of Non-Adap (Low-Adap). Algorithm Low-Adap considers the same sequence of tests as Non-Adap while (adaptively) skipping non-informative tests  based on observed outcomes. 
For the non-identifiable instance (WISER-ORG) we used the $O(d+ \min(c,r)+\log m + \log \frac 1\eps)$-approximation algorithms with both \textit{neighborhood} and \textit{clique} stopping criteria  (see  Appendix~\ref{app:non-id}). 
%We also consider three different stopping criteria: \textit{unique} stopping for perfectly identifiable instances, \textit{neighborhood} and \textit{clique} stopping (defined in Section~\ref{sec:non-id}) for WISER-ORG dataset.
The implementations of the adaptive and non-adaptive algorithms are available online.\footnote{https://github.com/FatemehNavidi/ODTN ;  https://github.com/sjia1/ODT-with-noisy-outcomes}   
\begin{table}[h!]
\footnotesize
\centering
\begin{tabular}{||c||c|c|c|c|c|c||}
\hline
\backslashbox[2.5cm]{Algorithm}{Data}&WISER-ID&Cl-0&Cl-5&Cl-10&Cl-20&Cl-30\\\hline\hline
{\bf Low-BND} & {\bf 7.994}&  {\bf 7.870} &{\bf 7.870} &{\bf 7.870} &{\bf 7.870}&{\bf 7.870}\\\hline
$\textrm{ODTN}_r$ & 8.357 & 7.910  & 7.927 & 7.915 & 7.962 & 8.000\\\hline
$\textrm{ODTN}_h$ & 9.707 & 7.910 & 7.979 & 8.211 & 8.671 & 8.729 \\\hline
Non-Adap & 11.568 & 9.731 & 9.831  & 9.941 & 9.996 & 10.204  \\\hline
Low-Adap & 9.152 & 8.619 & 8.517  &   8.777& 8.692&8.803 \\\hline
\end{tabular}
\vspace{2mm}
\caption{Cost of Different Algorithms for $\alpha = 0$ (Uniform Distribution).}
\label{tbl:alpha:0}
\end{table}

%%%%%%%%%%% dropped std-dev %%%%%%%%%%%%%%%%%
\ignore{
\begin{table}[h!]
\footnotesize
\centering
\begin{tabular}{||c||c|c|c|c|c|c||}
\hline
\backslashbox[2.5cm]{Algorithm}{Data}&WISER-ID&Cl-0&Cl-5&Cl-10&Cl-20&Cl-30\\\hline\hline
$\textrm{ODTN}_r$ & 0.008 & 0 & 0 & 0.002 & 0.003 & 0.006 \\\hline
$\textrm{ODTN}_h$ & 0.01 & 0 & 0 & 0 & 0.004 & 0.01 \\\hline
$\textrm{Non-Adap}$ &1.463 & 0.937 &1.047  &  1.092 & 1.056 &1.158  \\\hline
$\textrm{Low-Adap}$ & 0.0317 & 0.0685 & 0.0541 & 0.0760 & 0.0206 & 0.0550 \\\hline
\end{tabular}
\vspace{2mm}
\caption{Standard Deviation of Different Algorithms for $\alpha = 0$ (Uniform Distribution).}
\label{tbl:alpha:0:sd}
\end{table}
}
%%%%%%%%%%%%%%%%%%%%%%%%%%%%%%%%%%

\begin{table}[h!]
\footnotesize
\centering
\begin{tabular}{||c||c|c|c|c|c|c||}
\hline
\backslashbox[2.5cm]{Algorithm}{Data}&WISER-ID&Cl-0&Cl-5&Cl-10&Cl-20&Cl-30\\\hline\hline
Low-BND & 7.702&7.582&7.582&7.582&7.582&7.582\\\hline
$\textrm{ODTN}_r$ & 8.177 & 7.757 &7.780 & 7.789 & 7.831 & 7.900\\\hline
$\textrm{ODTN}_h$  & 9.306& 7.757 & 7.829 & 8.076 & 8.497 & 8.452\\\hline
Non-Adap & 11.998 &9.504 &9.500 & 9.694 &9.826 & 9.934 \\\hline
Low-Adap &8.096 &7.837& 7.565 &7.674 & 8.072 & 8.310 \\\hline
\end{tabular}
\caption{Cost of Different Algorithms for $\alpha = 0.5$.  {\new The (Low-adap, Cl-5) entry ($7.565$) is lower than Low-BND due to error in the sampling.}}
\label{tbl:alpha:0.5}
\end{table}

\begin{table}[h!]
\footnotesize
\centering
\begin{tabular}{||c||c|c|c|c|c|c||}
\hline
\backslashbox[2.5cm]{Algorithm}{Data}&WISER-ID&Cl-0&Cl-5&Cl-10&Cl-20&Cl-30\\\hline\hline
Low-BND &6.218 &6.136&6.136&6.136&6.136&6.136\\\hline
$\textrm{ODTN}_r$ & 7.367 & 6.998 & 7.121 & 7.150 & 7.299 & 7.357\\\hline
$\textrm{ODTN}_h$  & 8.566& 6.998 & 7.134 & 7.313 & 7.637  & 7.915\\\hline
Non-Adap &11.976&9.598  &9.672  &9.824  &10.159 &10.277  \\\hline
Low-Adap &9.072 &8.453 &8.344  &  8.609 &8.683 &  8.541\\\hline
\end{tabular}
\caption{Cost of Different Algorithms for $\alpha = 1$.}
\label{tbl:alpha:1}
\end{table}

\begin{table}[h!]
\footnotesize
\centering
\begin{tabular}{||c||c|c|c|c|c|c|c||}
\hline
\backslashbox[2.5cm]{Parameters}{Data}&WISER-ORG&WISER-ID&Cl-0&Cl-5&Cl-10&Cl-20&Cl-30\\\hline\hline
r &388& 245&0&5&7&12&13 \\\hline
Avg-r &50.46& 30.690&0 & 1.12 & 2.21 & 4.43 & 6.54\\\hline
h&61&45 & 0 & 3 & 6 & 8 & 8 \\\hline
Avg-h &9.51&9.39& 0 & 0.48 & 0.94 & 1.89 & 2.79 \\\hline
\end{tabular}
\vspace{2mm}
\caption{Maximum and Average Number of Stars per Hypothesis and per Test in Different Data sets.}
\label{tbl:param}
\end{table}

\begin{table}[h]
\footnotesize
\centering
\begin{tabular}{||c||c|c|c|c|c||}
\hline
Algorithm  & Neighborhood Stopping & Clique Stopping \\\hline\hline
$\textrm{ODTN}_r$ & 11.163   & 11.817   \\\hline
$\textrm{ODTN}_h$ & 11.908   & 12.506   \\\hline
Non-Adap & 16.995  & 21.281   \\\hline
Low-Adap & 16.983  & 20.559 \\\hline\end{tabular}
\caption{Algorithms on WISER-ORG dataset with Neighborhood and Clique Stopping for Uniform Distribution.}\label{tbl:non-id}
\end{table}

\noindent{\bf Results.}
Table~\ref{tbl:alpha:0}, Table~\ref{tbl:alpha:0.5} and Table~\ref{tbl:alpha:1} show the expected costs of different algorithms on all uniquely identifiable data sets when the parameter $\alpha$ in the distribution over hypothesis is $0, 0.5$ and $1$ correspondingly. 
These tables also report values of an information-theoretic lower bound (the entropy) on the optimal cost (Low-BND),  {\new estimated using $200$ \indep\ samples}. 
%We also report the sample standard deviation of all algorithms for uniform distribution in  Table~\ref{tbl:alpha:0:sd}.
As the approximation ratio of our algorithms depend on maximum number $c$ of unknowns per hypothesis and the maximum number $r$ of unknowns per test, we have also included these parameters as well as their average values in Table~\ref{tbl:param}. 
Table~\ref{tbl:non-id} summarizes the results on WISER-ORG with clique and neighborhood stopping criteria. 
We can see that $\textrm{ODTN}_r$ consistently outperforms the other algorithms and is very close to the information-theoretic lower bound. %Note that WISER-ORG dataset that is used to produce results in Table~\ref{tbl:non-id} has $m=414$ hypotheses and $d=54$ in its similarity graph, while WISER-ID in Table~\ref{tbl:alpha:0} is perfectly identifiable with $m=255$ hypotheses.

\acks{R. Ravi is supported in part by the U.S. Office of Naval
Research under award number N00014-21-1-2243 and the Air Force Office of Scientific Research under award number FA9550-23-1-0031. Viswanath Nagarajan is supported by NSF grants CMMI-1940766 and CCF-2006778.}

\bibliography{common_files/main}

\newpage
\appendix
\section{Proof of Proposition~\ref{prop:equivlence}: Reduction from ASRN to ASR}
\label{apdx:reduction}
% Recall that an \adap\ \alg\ for ASRN or ASR can be viewed as a \dec\ tree. 
It suffices to show that any feasible decision tree for the ASR instance $\mathcal{J}$ is also feasible for the ASRN instance $\mathcal{I}$ with the same objective and vice versa.
% If this is true, then by the definition of the prior \prb\ mass $\pi_{i,\omg}$ of each expanded \sce, we immediately deduce that the each \dec\ tree 
% implies that the cost of every decision these two \ins es are equivalent.

First, consider a \feas\ decision tree $\ho{T}$ for the ASR instance $\mathcal{J}$.
For any expanded scenario $(\ex) \in H$, let $P_\ex$ be the unique path traced in ${\ho{T}}$, and $S_\ex$ the elements selected along this path.
By the definition of a feasible \dec\ tree, at the last node (i.e., leaf) of $P_{\ex}$, we have $f_\ex(S_\ex)=1$, i.e., \[f_i(\{(e,\omega_e) : e\in S_\ex\})=1.\]
Therefore, $\ho{T}$ is a \feas\ \dec\ tree to $\I$.
% Now consider a \feas\ decision tree ${\ho{T}}$ for the ASRN instance ${\cal I}$. 
% \Sps\ the target \sce\ is $i\in [m]$ and the \elem-outcomes are given by $\omega\in \Omg^n$ (but still \unk\ to the \alg), which occurs with \prb\ $\pi_\ex = \op_i/|\Omega|^{s_i}$.
% Then, it is clear that the outcome from any element $e$ is $r_\ex(e)$ and hence the path traced in ${\ho{T}}$ is just $P_\ex$.
% Moreover, $f_i(\{(e,\omega_e) : e\in S_\ex\}) = 1$, which means we have $f_{i,\omega}(S_\ex) = 1$. 
% Taking an expectation over all $i$ and $\omega$, it follows that ${\ho{T}}$ is a feasible decision tree for ${\cal I}$ with cost at most that for instance ${\cal J}$.

Now, we consider the other direction. Let $\ho{T}'$ be a decision tree for the ASRN instance $\mathcal{I}$. 
\Sps\ the true scenario is $i\in [m]$ and the outcomes are given by a consistent vector $\omega\in \Omg^n$. 
Then, a unique path $P'_\ex$ is traced in $\ho{T}'$, whose \elem s we denote by $S'_\ex$.
Since $i$ is covered at the end of $P'_\ex$, we have $f_i(\{(e,\omega_e) : e\in S'_\ex\}) = 1$. 
Now view $\ho{T}'$ as a decision tree for the ASR instance $\mathcal{J}$. 
Then, the expanded scenario $(\ex)$ corresponds to a unique path $P'_\ex$, and therefore the elements $S'_\ex$ are selected. 
It follows that \[f_\ex(S'_\ex)= f_i(\{(e,\omega_e) : e\in S'_\ex\}) = 1,\] i.e., $(\ex)$ is covered at the end of $P'_\ex$. 
Therefore, $\ho{T}'$ is also a feasible tree for $\mathcal{J}$. \qed

\section{Details for the SFRN Problem (Section \ref{sec:nonadap})}
\label{apdx:nonadap}
Recall that the non-adaptive SFRN algorithm (Algorithm \ref{alg:non-adp}) involves two phases. 
In the first phase, we run the SFR algorithm using sampling to obtain estimates $\overline{G_E}(e)$ of the scores. 
If at some step, the maximum sampled score is ``too low'' then we go to the second phase where we perform all remaining \elem s in an \arb\ order. 
The number of samples used to obtain each estimate is polynomial in $m,n,\eps^{-1}$, so the overall runtime is polynomial. 

\noindent{\bf Pre-processing.} We first show that by losing an $O(1)$-factor in \apxn\ ratio, we may assume that $\pi_i\geq n^{-2}$ for all $i\in [m]$.  
Let $A=\{i\in [m]:\pi_i\leq n^{-2}\}$, then $\sum_i \pi_i \leq n^{-2}\cdot n\leq n^{-1}$. 
Replace all \sce s in $A$ with a single dummy \sce\ ``$0$'' with $\pi_0=\sum_{i\in A}\pi_i$, and define $f_0$ to be any $f_i$ where $i\in A$.
By our assumption that each $f_i$ must be covered irrespective of the noisy outcomes, it holds that $f_{i,\omg}([n])=1$ for each $\omega\in \Omega(i)$, and hence the cover time is at most $n$.
Thus, for any \perm\ $\sigma$, the expected cover time of the old and new \ins\ differ by at most $O(n^{-1} \cdot n)=O(1)$. 
Therefore, the cover time of any \xulie\ of \elem s differs by only $O(1)$ in this new instance (where we removed the \sce s with tiny prior densities) and the original \ins s. 

% We now present the formal proof of Theorem~\ref{thm:non-adp-sfrn}, with proofs of the lemmas deferred to Appendix~\ref{apdx:nonadap}. 
To analyze our randomized \alg, we need the following sampling lemma, which follows from the standard Chernoff bound.
\begin{lemma}[Concentration Bound]\label{lem:chernoff}
Let $X$ be a bounded random variable with $\E X \ge m^{-2} n^{-4}\eps$ and $X\in [0,1]$ a.s. 
Denote by $\bar{X}$ the average of $m^3 n^4 \eps^{-1}$ many independent samples of $X$. 
Then, \[\Pr\left[\bar{X} \notin \lb[ \frac12 \E X,2\E X\rb] \right] \leq e^{-\Omega(m)}\]
\end{lemma}
\proof Let $X_1,...,X_N$ be i.i.d. samples of random variable where $N=m^3 n^4 \eps^{-1}$ is the number of samples. 
Letting $Y=\sum_{i\in [N]} X_i$, Chernoff's  \ineq\ implies for any $\delta\in (0,1)$,  
$$\Pr\lb(Y \notin [(1-\delta) \E Y,(1+\delta) \E Y] \rb)\le \exp\lb(-\frac{\delta^2}2 \cdot \E Y\rb).$$
The claim follows by setting $\delta=\frac12$ and using the assumption that  \[\E Y = N \cdot \E X_1 = \Omega(m).\eqno\qed\] 

The next lemma shows that sampling does find an approximate maximizer unless the score is very small, and also bounds the {\it failure} probability.

\bdefn[Failure] Consider any \iter\ in Algorithm \ref{alg:non-adp} with $S =\max_{e\in [n]} {G_E}(e)$ and $\bar{S}=\max_{e\in [n]} \overline{G_E}(e)$
with $\overline{G_E}(e^*)=\bar{S}$. 
We say that this step is a {\bf failure} if either (i) $\bar{S} < \frac14 m^{-2}n^{-4}\eps$ and $S\ge \frac12 m^{-2}n^{-4}\eps$, or (ii) $\bar{S} \ge \frac14 m^{-2}n^{-4}\eps$ and $G_E(e^*)< \frac{S}4$.
\edefn

\begin{lemma}[Failure Probability Is Low]\label{lem:phase1-non-adp} 
The probability of failure is $e^{-\Omega(m)}$.
\end{lemma}
\proof We will consider the two types of failure separately. 
For the first type, suppose $S \ge  \frac12 m^{-2}n^{-4}\eps$. 
Applying Lemma~\ref{lem:chernoff} on the \elem\ $e\in [n]$ with $G_E(e)=S$, we obtain $$\Pr\lb[\bar{S} < \frac14 m^{-2}n^{-4}\eps\rb] \le \Pr\lb[ \overline{G_E}(e) < \frac14 m^{-2}n^{-4}\eps\rb] \le e^{-\Omega(m)}.$$ 
So the probability of the first type of failure is at most $e^{-\Omega(m)}$.
For the second type of failure, we consider two cases.

{\bf Case 1:} \Sps\ $S<\frac18 m^{-2}n^{-4}\eps$.
For any $e\in [n]$ we have $G_E(e)\le S< \frac18 m^{-2}n^{-4}\eps$. 
Note that $\overline{G_E}(e)$ is the average of $N$ independent draws, each with mean $G_E(e)$. We now upper bound the probability of the event ${\cal B}_e$ that $\overline{G_E}(e)\ge \frac14 m^{-2}n^{-4}\eps$.  
We first artificially increase each sample mean to $\frac18 m^{-2}n^{-4}\eps$: note that this only increases the probability of the event ${\cal B}_e$. 
 Now, using Lemma~\ref{lem:chernoff} we obtain $\Pr[{\cal B}_e]\le e^{-\Omega(m)}$. 
 By a union bound, it follows that $\Pr[\bar{S} \ge \frac14 m^{-2}n^{-4}\eps]\le \sum_{e\in [n]} \Pr[{\cal B}_e] \le e^{-\Omega(m)}$. 

{\bf Case 2:} \Sps\ $S \ge \frac18 m^{-2}n^{-4}\eps$. 
%As argued above, $\bar{S}\ge   \frac14 m^{-5}$ with probability $1-e^{-\Omega(m)}$. 
Consider now any $e\in U $ with $G_E(e)< S/4$. 
By Lemma~\ref{lem:chernoff} (artificially increasing $G_E(e)$ to $S/4$ if needed), it follows that $\Pr[\overline{G_E}(e) > S/2] \le e^{-\Omega(m)}$. 
Now consider the \elem\ $e'$ with $G_E(e')=S$. Again, by Lemma~\ref{lem:chernoff}, it follows that $\Pr[\overline{G_E}(e') \le S/2] \le e^{-\Omega(m)}$. 
This means that \elem\ $e^*$ has $\overline{G_E}(e^*) \ge \overline{G_E}(e')> S/2$ and $G_E(e^*)\ge S/4$ with probability $1-e^{-\Omega(m)}$. 
In other words, assuming $S \ge \frac18 m^{-2}n^{-4}\eps$, the probability that $G_E(e^*)<  S/4$ is at most $e^{-\Omega(m)}$.

Adding the probabilities over all possibilities for failures, the lemma follows.  \qed

Based on Lemma~\ref{lem:phase1-non-adp}, in the remaining analysis, we condition on the event that our algorithm never encounters failures, which occurs with probability $1-e^{-\Omega(m)}$. 
To conclude the proof, we need the following key lemma which essentially states that if the score of the greediest \elem\ is low, then the \elem s selected so far suffices to cover {\it all} \sce s with high \prb, and therefore the ordering of the remaining \elem s does not matter much.
\begin{lemma}[Handling Small Greedy Score]\label{lem:phase2-non-adp}
Assume that there are no failures. 
Consider the end of phase 1 in our algorithm, i.e., the first step with $\overline{G_E}(e^*) < \frac14 m^{-2}n^{-4}\eps$. 
Then, the probability that the realized scenario is not covered is at most $m^{-2}$.  
\end{lemma}
\proof Let $E$ denote the \elem s chosen so far and $p$ the probability that $E$ does {\em not} cover the realized scenario-copy of $H$, formally,  
\[ p= \Pr_{(i,\omg)\in H}(f_{i,\omg}(E)<1) = \sum_{i=1}^m \pi_i\cdot \Pr_{\omg\in \Omega(i)}(f_{i,\omg}(E)<1).\]

It follows that there is some $i$ with $\Pr_{\omg\in \Omega(i)}(f_{i,\omg}(E)<1)\ge p$. 
By definition of separability, if $f_{i,\omg}(E)<1$ then $f_{i,\omg}(E)\le 1-\eps$. Thus, 
$$\sum_{\omg\in \Omega(i)} \pi_{i,\omg}f_{i,\omg}(E) \leq \sum_{\omg: f_{i,\omg}(E)=1} \pi_{i,\omg}\cdot 1 + \sum_{\omg: f_{i,\omg}(E)< 1}\pi_{i,\omg}\cdot f_{i,\omg}(E) \leq (1-\eps p)\pi_i .$$
On the other hand, taking all the elements, we have $f_{i,\omg}([n])=1$ for all $\omg \in \Omega(i)$. 
Thus, 
$$\sum_{\omg\in \Omg(i)} \pi_{i,\omg}f_{i,\omg}([n]) = \sum_{\omg\in \Omega(i)} \pi_{i,\omg} =\pi_i.$$
Taking the difference of the above two inequalities, we have
$$\sum_{\omg\in \Omg(i)} \pi_{i,\omg}\cdot (f_{i,\omg}([n])-f_{i,\omg}(E))\geq  \pi_i \cdot \eps p.$$
Consider function $g(S) := \sum_{\omg\in \Omg(i)} \pi_{i,\omg}\cdot \left( f_{i,\omg}(S\cup E) - f_{i,\omg}(E)\right)$ for $S\sse [n]$, which is also submodular. From the above, we have $g([n]) \ge \pi_i \cdot \eps p$. 
Using the submodularity of $g$,
$$\max_{e\in [n]} g(\{e\}) \ge \frac{\eps p \pi_i }{n} \implies \exists \tilde  e\in [n] \,:\, \sum_{\omg\in\Omega(i)} \pi_{i,\omg}\cdot \left( f_{i,\omg}(E\cup \{\tilde e\}) - f_{i,\omg}(E)\right) \geq \frac{\eps p \pi_i }{n} .$$
It follows that $G_E(\tilde{e})\geq \frac{\eps p \pi_i }{n} \ge n^{-3} \eps p$, where we used $\min_i \pi_i \ge n^{-2}$. Now, suppose for a contradiction that $p\ge m^{-2}$. 
Since there is no failure and $G_E(\tilde{e})\geq n^{-3} m^{-2} \eps\geq \frac{1}{4} n^{-4} m^{-2}\eps$, by case (ii) of Lemma~\ref{lem:phase1-non-adp} , we deduce that $\overline{G_E}({e^*})\ge \frac14 m^{-2}n^{-4}$, a contradiction.
\qed

The above is essentially a consequence of the submodularity of the target \func s.
\Sps\ for \contra\ that there is a \sce\ $i$ that, with at least $m^{-2}$ \prb\ over the random outcomes, remains {\it uncovered} by the currently selected \elem s. 
Recall that according to our feasibility assumption, if all \elem s were selected, then $f_i$ is covered with \prb\ $1$. 
Therefore, by submodularity, there exists an individual \elem\ $\tilde e$ whose inclusion brings more coverage than the \avg\ coverage over all \elem s in $[n]$, and therefore $\tilde e$ has a ``high'' score.

\noindent{\bf Proof of Theorem~\ref{thm:non-adp-sfrn}.} Assume that there are no failures. 
We proceed by bounding the expected costs (number of \elem s) from phases 1 and 2 separately. 
By Lemma~\ref{lem:phase1-non-adp}, the element chosen in each step of phase 1 is a 4-approximate maximizer (see case (ii) failure) of the score used in the SFR algorithm. 
Thus, by Theorem~\ref{thm:sfr-apx}, the expected cost in phase 1 is $O(\log m)$ times the optimum. 
On the other hand, by Lemma~\ref{lem:phase2-non-adp} the probability of performing phase 2 is at most $e^{-\Omg(m)}$. 
As there are at most $n$ \elem s in phase 2, the expected cost is only $O(1)$. 
Therefore, \Alg~\ref{alg:non-adp} is an $O(\log m)$-approximation algorithm for the SFRN problem. \qed

\section{Efficient Implementation of \Alg\ \ref{alg:asr}}
\label{apdx:computation_score}
As we recall, it was not clear why the score function in \Alg\ \ref{alg:asr} can be efficiently computed. 
In this section, we explain why this \alg\ can be implemented in polynomial time. 

\subsection{Computing the First Term in $\mathrm{Score}_c$.} 
Recall that $H_i$ is the set of all expanded \sce s for $i$.
Since each $(\ex)\in H_i$ has an equal share $\pi_\ex = |\Omega|^{-c_i}\pi_i$ of prior \prb\ mass the (original) \sce\ $i\in [m]$, computing the first term in ${\rm Score}_c$ reduces to maintaining the {\em number} $n_i = |H_i \cap H'|$ of consistent copies of $i$.
We observe that $n_i$ can be easily updated in each iteration. 
In fact, \sps\ outcome $o\in \Omg$ is observed when selecting \elem\ $e$. 
We consider how $H'\cap H_i$ changes after selecting in the following three cases.
\benum
\item if $r_i(e)\notin \{\star, o\}$, then none of $i$'s expanded \sce s would remain in $H'$, so $n_i$ becomes 0, 
\item if $r_i(e)=o$, then all of $i$'s expanded \sce s would remain in $H'$, so $n_i$ remains the same,
\item if $r_i(e)= \star$, then only those $(i,\omg)$ with $\omg(e) = o$ will remain, and so
$n_i$ shrinks by an $|\Omg|$ factor.
\eenum

As $n_i$'s can be easily updated, we are also able to compute the first term in $\mathrm{Score}_c$ efficiently.
Indeed, for any \elem\ $e$ (that is not yet selected), we can implicitly describe the set $L_e(H')$ as follows. 
Note that for any outcome $o\in \Omg$, 
$$| \{(\ex)\in H' : r_\ex(e) = o\} | = \sum_{i\in [m] : r_i(e)=o} n_i \,+\, \frac 1 {|\Omg|} \sum_{i\in [m] : r_i (e)=\star} n_i,$$ 
so the largest cardinality set $B_e(H')$ can be easily determined using $n_i$'s.
In fact, let $b$ be the outcome corresponding to $B_e(H')$.
% Moreover, its 
% weight can be calculated as
Then, 
\[\pi\left(L_e\left(H'\right)\right)
%:=\sum_{(\ex)\in L_e(H')}\pi_\ex
=\sum_{i\in [m]: r_i(e)\notin \{b, \star\}} \frac{\pi_i}{|\Omega|^{c_i}} \cdot n_i \,+\, \frac{|\Omg|-1}{|\Omg|} \sum_{i\in [m]: r_i(e)=\star} \frac{\pi_i}{|\Omega|^{c_i}} \cdot n_i.\]
% To suppress notation, we use $\pi(S)=$ for any subset $S\sse H$. 

\subsection{Computing the Second Term in $\mathrm{Score}_c$}
The second term in $\mathrm{Score}_c$ involves summing over exponentially many terms, so a naive implementation is inefficient. 
Instead, we will rewrite this summation as an {\it expectation} that can be calculated in polynomial time.

% \Sps\ $e\notin E$ and for any $o\in \Omg$, define $r_e^{-1}(o) = \{i\in [m]: r_i(e) = o\}$.

We introduce some notation before formally stating this equivalence.
\Sps\ the \alg\ selected a subset $E$ of \elem s, and observed outcomes $\{\nu_e\}_{e\in E}$.
We overload  notation slightly and use  $f(\nu_E) := f\big(\{(e,\nu_e) : e\in E\}\big)$ for any function $f$ defined on $2^{[m]\times \Omg}$.
%and consider the set of {\it surviving} (original) \sce s 
% \[S = \{i\in [m] : H_i \cap H' \ne \emptyset\} = \{i\in [m] : n_i\ge 1\}.\]
For each \sce\ $i\in [m]$, let $p_i = n_i \cdot \frac{\pi_i}{|\Omega|^{c_i}}$
be the total \prb\ mass of the surviving expanded \sce s for $i$.\footnote[2]{One may easily verify via the Bayesian rule that $p_i/p([m])$ is indeed the posterior probability of scenario  $i\in [m]$, given the previously observed outcomes.}
% For ease of notation, we will use the shorthand 
% $p(A):= \sum_{i\in A} p_i$ for any subset $A\sse [m]$ of original \sce s.
Finally, for any element $e$ and scenario $i$, let $\ho{E}_{i,\nu_e}$ be the expectation over the outcome $\nu_e$ of \elem\ $e$ conditional on $i$ being the realized \sce.
We can then rewrite %(Appendix~\ref{apdx:lemma5}) 
the second term in $\mathrm{Score}_c$ as follows. 
\begin{lemma}[Reformulation of the Greedy Score]
\label{lem:aug28}
For each $i\in [m]$, and $e\notin E$,
\begin{align}
\label{eq:asr-h-compact}
\sum\limits_{(\ex)\in H'} \pi_{\ex} \cdot \frac{f_{\ex}(e\cup E) - f_{\ex}(E)}{1-f_{\ex}(E)}
= \sum_{i\in [m]} p_i \cdot \frac{\ho{E}_{i,\nu_e} [f_i(\nu_E \cup \{ \nu_e \}) - f_i(  \nu_E)]}{1-f_i(\nu_E)}
\end{align}
\end{lemma}
\proof By decomposing the summation in the left hand side of \eqref{eq:asr-score} as $H' = \cup_i H'\cap H_i$, and noticing that $f_{i,\omg}(E) = f_i(\nu_E)$, 
the problem reduces to showing that for each $i\in [m]$,
\[\sum_{(i,\omg)\in H'\cap H_i} \pi_{i,\omg} \cdot \big(f_{\ex}(e\cup E) - f_{\ex}(E)\big) = p_i\cdot \ho{E}_{i,\nu_e} [f_i(\nu_E \cup \{  \nu_e \}) - f_i( \nu_E)].\]
Recall that $p_i = n_i \cdot \frac{\pi_i}{|\Omega|^{c_i}}$ and $\pi_{(\ex)} = \frac {\pi_i}{|\Omega|^{c_i}}$, the above simplifies to
\[\frac 1{n_i} \sum_{(i,\omg) \in H'\cap H_i}  \big(f_{\ex}(e\cup E) - f_{\ex}(E)\big) 
= \ho{E}_{i,\nu_e} [f_i(\nu_E \cup \{  \nu_e \}) - f_i( \nu_E)].\]
Note that $n_i = |H'\cap H_i|$, so the above is equivalent to
\begin{align}\label{eq:aug30}
\frac 1{n_i} \sum_{(i,\omg) \in H'\cap H_i} f_{\ex}(e\cup E) 
= \ho{E}_{i,\nu_e} [f_i\big (\nu_E \cup \{\nu_e \}\big)].
\end{align}
It is \strfwd\ to verify that the above by considering the following two cases. 

{\bf Case 1:}  If $r_i(e) = \nu_e \in \Omg \setminus \{*\}$, then the outcome $\nu_e$ is \dtmnstc\ conditional on \sce\ $i$, and so is $f_i\big(\nu_E \cup \{ \nu_e \}\big)$, the value of $f_i$ after selecting $e$.
On the left-hand side, for every $\omg \in H_i$, by definition of $H_i$ it holds $\nu_e = \omg_e$, and hence $f_{\ex}(e\cup E) = f_i(\nu_E \cup \{\nu_e\}$ for {\it every} $(\ex)\in H_i$.
Therefore all terms in the summation are equal to $f_i(\nu_E \cup \{\nu_e\}$
and hence \eqref{eq:aug30} holds.

{\bf Case 2:} If $r_i(e) = \star$, then each outcome $o\in \Omg$ occurs with equal probabilities, thus we may rewrite the right hand side as 
\begin{align*}
\ho{E}_{i,\nu_e} [f_i\lb (\nu_E \cup \{\nu_e \}\rb)] = \sum_{o\in \Omg} \ho{P}_i[\nu_e = o]\cdot f_i\lb(\nu_E \cup \{ \nu_e\}\rb)  = \frac 1{|\Omg|}\sum_{o\in \Omg} f_i\lb(\nu_E \cup \{(e, o)\}\rb).
\end{align*}
To analyze the other side, note that by the definition of $H_i$ and $H'$, there are equally many expanded \sce s $(i,\omg)$ in $H'\cap H_i$ with $\omg_e = o$ for each outcome $o\in \Omg$.
Thus, we can rewrite the left hand side as 
\begin{align*}
\frac 1{n_i} \sum_{(i,\omg) \in H'\cap H_i} f_{\ex}(e\cup E) 
&= 
\frac 1{n_i} \sum_{o\in \Omg} \sum_{\substack{(i,\omg) \in H'\cap H_i,\\ \omg_e = o}} 
 f_{\ex}(e\cup E) \\
& = \frac 1{n_i} \sum_{o\in \Omg} \frac{n_i}{|\Omg|}
f_{\ex}(e\cup E) \\
& = \frac 1{|\Omg|}
\sum_{o\in \Omg} f_i\big (\nu_E \cup \{(e, o)\}\big),
\end{align*}
which matches the right hand side of \eqref{eq:aug30} and completes the proof. \qed 

This lemma suggests the following efficient implementation of \Alg~\ref{alg:asr}.
For each $i$, compute and maintain $p_i$ using $n_i$.
%which we explained how to maintain previously.
To find the expectation in the numerator, note that if $r_i(e)\neq \star$, then $\nu_e$ is \dtmnstc\ and hence it is \strfwd\ to find this expectation.
In the other case, if $r_i(e) = \star$, recalling that the outcome is \unif\ over $\Omg$, we may simply evaluate $f_i(\nu_E \cup \{(e, o)\}) - f_i(\nu_E)$ for each $o\in \Omg$ and take the \avg, since the noisy outcome is \unif ly distributed over $\Omg$.

\section{Analysis of the ASRN Problem (Section \ref{sec:few_star})}
This section is dedicated to presenting the details of how we establish our results for the adaptive SFRN problem, mainly Theorem \ref{thm:asrn-c} and Theorem \ref{thm:odtn-r}.

\subsection{Application of \Alg~\ref{alg:asr} and \Alg~\ref{alg:asr_*} to ODTN.} \label{app:odtn-score}
%\vnote{We should give the closed form for the 2nd algo also.}
For concreteness, we provide a closed-form formula for $\mathrm{Score}_c$ and $\mathrm{Score}_r$ in the ODTN problem using Lemma~\ref{lem:aug28}, which were used in our experiments for ODTN. In \S\ref{subsec:asr-noise}, we formulated ODTN as an ASRN instance. Recall 
that the outcomes $\Omega=\{+1,-1\}$, and the submodular function $f$ (associated with each hypothesis $i$) measures the proportion of \hypos\ eliminated after observing the outcomes of a subset of tests. 

As in \S\ref{sec:few_star}, at any point in Algorithm~\ref{alg:asr} or \ref{alg:asr_*}, after selecting  set $E$ of tests, let $\nu_E: E \rar \pm 1$ denote their  outcomes. For each hypothesis $i\in [m]$, let $n_i$ denote  the number of surviving expanded-scenarios of $i$. Also, for each hypothesis $i$, let $p_i$ denote the total probability mass of the surviving expanded-scenarios of $i$.  For any $S\sse [m]$, we use the shorthand $p(S)=\sum_{i\in S} p_i$. Finally, let $A\sse [m]$ denote the compatible hypotheses based on the observed outcomes $\nu_E$ (these are all the hypotheses $i$ with $n_i>0$). 
Then, $f(\nu_E) = \frac{m-|A|}{m-1}$. Moreover, for any new test/element $T$, 
$$f(\nu_E\cup \{\nu_T\}) =  \left\{\begin{array}{ll}
   \frac{m-|A|+|A\cap T^-|}{m-1}  & \mbox{ if }\nu_T=+1 \\
     \frac{m-|A|+|A\cap T^+|}{m-1} & \mbox{ if }\nu_T=-1
\end{array}\right. .$$
Recall that  $T^+$, $T^-$  and $T^*$ denote the hypotheses with $+1$, $-1$ and  $*$ outcomes for test $T$. So,
$$\frac{f(\nu_E\cup \{\nu_T\}) -f(\nu_E)}{1-f(\nu_E)}=  \left\{\begin{array}{ll}
   \frac{ |A\cap T^-|}{|A|-1}  & \mbox{ if }\nu_T=+1 \\
     \frac{|A\cap T^+|}{|A|-1} & \mbox{ if }\nu_T=-1
\end{array}\right. .$$

 It is then \strfwd\ to verify the following.
% Consider the ODTN problem with \hypos\ space $[m]$ and outcome space $\Omg$.\iter, $A\subseteq [m]$ is the subset of \hypos\ consistent with all outcomes observed, and $p=(p_i)_{i\in [m]}$ is the \prb\ weights of the original \hypos. where\[\pi\left(R_T \left(H'\right)\right) =\sum_{o\in \Omg\bs b_T} \sum_{i\in T^o} \frac{\pi_i}{2^{c_i}} \cdot n_i \,+\, \frac{|\Omg|-1}{|\Omg|} \sum_{i\in T^*} \frac{\pi_i}{2^{c_i}} \cdot n_i.\]
\begin{proposition}
Consider implementing \Alg~\ref{alg:asr} on an ODTN \ins. 
\Sps\ after selecting tests $E$, the expanded-scenarios $H'$ (and original scenarios $A$) are compatible with the parameters described above. For any   test $T$, if $b_T\in\{+1,-1\}$ is the outcome \corres\ to $B_T(H')$  then the second term in $\mathrm{Score}_c (T;E,H')$ and $\mathrm{Score}_r (T;E,H')$ is:
\begin{align*}\label{eq:asr-h-compact}
\left( \frac{ |A\cap T^-|}{|A|-1}   + \frac{ |A\cap T^+|}{|A|-1}  \right) \cdot \frac{p\left(A\cap T^* \right)}2  + \frac{ |A\cap T^-|}{|A|-1}   \cdot p\left(A\cap T^+ \right)+ \frac{ |A\cap T^+|}{|A|-1}   \cdot p\left(A\cap T^- \right).
\end{align*}
\end{proposition}
The above expression has a natural interpretation for ODTN: conditioned on the outcomes $\nu_E$ so far, it is the expected number of newly eliminated hypotheses due to test $T$ (normalized by $|A|-1$).

The first term of the score $\pi\left(L_T\left(H'\right)\right)$ or $\pi\left(R_T\left(H'\right)\right)$ is calculated as for the general ASRN problem.
Finally, observe  that for  the \submod\ \func s used for  ODTN, the separation \pmt\ is $\eps = \frac 1{m-1}$. So, by Theorem~\ref{thm:asrn} we immediately obtain a polynomial time $O(\min(r,c)+ \log m)$-\apxn\ for ODTN.

\ignore{And similarly, we can obtain a closed form expression for $\mathrm{Score}_r$ as defined in \Alg~\ref{alg:asr_*}.
\begin{proposition}
Considering implementing \Alg~\ref{alg:asr_*} on an ODTN \ins.
\Sps\ at some \iter, $S\subseteq [m]$ is the subset of \hypos\ consistent with all outcomes observed, and $p=(p_i)_{i\in [m]}$ is the \prb\ weights of the original \hypos. 
For any test $T$, let $\bar o\in \Omg$ be the outcome \corres\ to $C_T(S)$ as defined in \Alg~\ref{alg:asr_*}, then
\begin{align*}\label{eq:asr-h-compact}
\mathrm{Score}_c (T;E,H') = \pi\left(R_T\left(H'\right)\right) + \frac1{|\Omg|} \frac{|\cup_{o\in \Omg} T^o \cap A|}{m-1}\cdot p\left(T^* \right) +
\sum_{o\in \Omg} \frac{|\cup_{o'\in \Omg\bs \{o\}}T^{o'} \cap A|}{m-1}\cdot p\left(T^o \right),
\end{align*}
where
\[\pi\left(R_T \left(H'\right)\right)
=\sum_{i\in T^{\bar o}} \frac{\pi_i}{2^{c_i}} \cdot n_i \,+\, \frac1{|\Omg|} \sum_{i\in T^*} \frac{\pi_i}{2^{c_i}} \cdot n_i.\]
\end{proposition}
These expressions can obviously be computed in polynomial time, since $p$ can be computed from $n_i$'s which we showed how to maintain efficiently.  
Note that by definition of the \submod\ \func s in \eqref{eq:def_fi}, the separation \pmt\ is $\eps = \frac 1{m-1}$, we immediately obtain a polynomial time $O(\min(r,c)+ \log m)$-\apxn\ for ODTN.
}

\subsection{Proof of Theorem~\ref{thm:odtn-r}}
\label{apdx:r} 
\begin{algorithm}\begin{algorithmic}[1]
\State Initialize $E \leftarrow \emptyset, H' \leftarrow H$
\While {$H'\neq \emptyset$} 
\State $S\gets \{i\in [m]\,:\, H_i\cap H' \ne \emptyset\}$\Comment{Consistent original scenarios}
\State For $e\in [n]$, let $U_e(S) = \{i\in S: r_i(e) = *\}$ and  $C_e(S)$ be the largest cardinality set among
$$\{i\in S: r_i(e) = o\}, \quad \forall o\in \Omg,$$
and let $o_e(S)\in \Omega$ be the outcome corresponding to $C_e(S)$.
\State For each $e\in [n]$, let 
$$\overline{R_e}(H') = \{(\ex)\in H': i \in C_e(S)\} \bigcup \{(j,o_e(S))\in  H' : j\in U_e(S)\},$$
be  those expanded-scenarios that have outcome $o_e(S)$ for element $e$, and   $R_e(H'):= H'\setminus \overline{R_e}(H')$.
\State Select element $e\in [n]\setminus E$ that maximizes
\begin{equation} \label{eq:asr-r-score}
\mathrm{Score}_r(e,E,H') = \pi\big(R_e(H')\big) \,\,+\,\, \sum\limits_{(\ex)\in H', f_{\ex}(E)<1} \pi_{\ex} \cdot \frac{f_{\ex}(e\cup E)-f_{\ex}(E)}{1-f_{\ex}(E)}
\end{equation}
\State Observe outcome $o$
\State $H'\lar \{(\ex)\in H': r_\ex(e)=o\mbox{ and } f_{\ex}(E\cup e)<1\}$
\Comment{Update the  (expanded) scenarios}  
% scenarios  from $H'$ based on the feedback from $e$
\State $E\leftarrow E\cup \{e\}$
\EndWhile
\caption{Modified algorithm for ASR instance \J. \label{alg:asr_*}}
\end{algorithmic}
\end{algorithm}

The proof is  similar to the analysis in  \cite{navidi2016adaptive}. 
With some foresight, set $\alpha := 15(r+\log m)$.
Write Algorithm~\ref{alg:asr_*} as ALG and let OPT be the optimal adaptive policy. 
It will be  convenient to view ALG and OPT as \dec\ trees where each node represents the ``state'' of the policy. Nodes in the decision tree are labelled by elements (that are selected at the corresponding state) and branches out of each node are labelled by the outcome observed at that point. At any state, we use $E$ to denote the previously selected elements and $H'\sse M$ to denote the {\em expanded-scenarios} that are (i) compatible with the  outcomes observed so far and (ii) uncovered.
\Sps\ at some \iter, \elem s $E$ are selected and outcomes $\nu_E$ are observed, then a \sce\ $i$ is said to be {\it covered} if $f_i(E\cup \nu_E) = 1$, and {\it uncovered} \ow.

For ease of presentation, we use the phrase ``at time $t$'' to mean ``after selecting $t$ \elem s''. Note that the cost incurred until time $t$ is exactly $t$.
The key step is to show 
\begin{equation}\label{eq:ASR-analysis}
a_k \le 0.2 a_{k-1} + 3 y_k, \qquad  \mbox{ for all }k\ge 1,
\end{equation}
where
\begin{itemize}
\item $A_k\sse M$ is the set of uncovered expanded \sce s in ALG at time $\alpha\cdot 2^k$ and $a_k=p(A_k)$ is their total probability,
\item $Y_k$ is the set of uncovered \sce s in OPT at time $2^{k-1}$, and $y_k=p(Y_k)$ is the total probability of these \sce s.
\end{itemize}

As shown in Section 2 of \cite{navidi2016adaptive}, %by simple arithmetic manipulations, 
\eqref{eq:ASR-analysis}
implies that Algorithm~\ref{alg:asr_*} is an $O(\alpha)$-approximation
and hence Theorem~\ref{thm:odtn-r} follows. 
To prove \eqref{eq:ASR-analysis}, we consider the total score collected by ALG between iterations $\alpha 2^{k-1}$ and $\alpha 2^k$, formally given by
{\small \begin{align}\label{eq:gain_*}
Z  &:= \sum\limits_{t>\alpha 2^{k-1}}^{\alpha 2^k}\enskip\sum\limits_{(E,H')\in V(t)} \,\, \max_{e\in [n]\setminus E}
\left(\sum_{(\ex)\in R_e(H')}\pi_{\ex} \,\,+\,\, \sum\limits_{(\ex)\in H'} \pi_{\ex} \cdot \frac{f_{\ex}(e\cup E)-f_{\ex}(E)}{1-f_{\ex}(E)}\right)
\end{align}} 
where $V(t)$ denotes the set of states $(E,H')$ that occur at time  $t$ in the \dec\ tree ALG. We note that all the expanded-scenarios seen in states of $V(t)$
are contained in $A_{k-1}$. 

\def\st{\mathsf{Stem}_k(H')}

Consider any state $(E,H')$ at time $t$ in the algorithm. Recall that $H'$ are the expanded-scenarios and let $S\sse [m]$ denote the original scenarios in $H'$. Let $T_{H'}(k)$ denote the subtree of OPT that corresponds to paths traced by expanded scenarios in $H'$ up to time $2^{k-1}$.   
Note that each node (labeled by any element $e\in [n]$) in $T_H(k)$ has at most $|\Omega|$ outgoing branches and  one of them 
 corresponds to  the outcome $o_e(S)$ defined in Algorithm~\ref{alg:asr_*}. We  define  $\st$ to be the path in $T_{H'}(k)$ that at each node (labeled $e$) follows the  $o_e(S)$ branch. 
We also use $\st\sse [n]\times  \Omega$ to denote the observed element-outcome pairs  on this path.

\begin{definition}\label{defn:type} Each state $(E,H')$  is exactly one of the following types:
\begin{itemize}
\item {\bf bad} if the probability of  uncovered scenarios in $H'$ at the end of $\st$ is at least $\frac{\Pr(H')}{3}$.
\item {\bf okay} if it is not bad and  
$\Pr(\cup_{e\in \st} \, R_e(H'))$  is at least $\frac{\Pr(H')}{3}$.
\item {\bf good} if it is neither bad nor okay and the probability of scenarios in $H'$ that get covered by $\st$  is at least $\frac{\Pr(H')}{3}$.
\end{itemize}
\end{definition}
Crucially, this categorization of states is well defined. Indeed,  each expanded-scenario in $H'$ is {\bf (i)} uncovered at the end of $\st$, or {\bf (ii)} in $R_e(H')$ for some $e\in \st$, or {\bf (iii)} covered by some prefix of $\st$, i.e. the function value reaches $1$ on $\st$. So the total probability of the scenarios in one of these $3$ categories must be at  least $\frac{\Pr(H)}{3}$. %Therefore each state $(E,H)$ is exactly one of these three types. 

In the next two lemmas, we will show a lower bound (Lemma~\ref{lem:GLB}) and an upper bound (Lemma~\ref{lem:GUB}) for $Z$ in terms of $a_k$ and $y_k$, which together imply \eqref{eq:ASR-analysis} and complete the proof.

\begin{lemma} \label{lem:GLB} 
For any $k\ge 1$, it holds $Z\geq \alpha\cdot(a_k - 3y_k)/3$. 
\end{lemma}
%\vnote{Should add the proof as there was an issue with the earlier algorithm.  }
\proof
The proof of this lower bound is identical to that of Lemma~3 in \cite{navidi2016adaptive} for noiseless-ASR. The only  difference is that we use the scenario-subset  $R_e(H')\sse H'$ instead of subset  ``$L_e(H)\sse H$'' in the  analysis of \cite{navidi2016adaptive}.
%The rest of the analysis is identical to that of  Lemma~3 in \cite{navidi2016adaptive}.
\qed

%Proposition~2 in \cite{navidi2016adaptive} continues to hold. So, the total probability of bad states in $V(t)$ is at most $3y_k$. 

\begin{lemma}\label{lem:GUB} 
For any $k\ge 1$, $Z \leq a_{k-1} \cdot(1+\ln{\frac1\epsilon}+ r+\log{m})$.
\end{lemma}
\proof
This proof is analogous to that of Lemma~4 in \cite{navidi2016adaptive} but requires new ideas, as detailed below.
Our proof splits into two steps.
We first rewrite $Z$ by interchanging its  double summation:  the outer layer is now over the $A_{k-1}$ (instead of  times between $\alpha 2^{k-1}$ to $\alpha 2^k$ as in the original definition of $Z$).
Then for each fixed $(\ex)\in A_{k-1}$, we will upper bound the inner summation using the assumption that there are at most $r$ original \sce s with $r_i(e) = \star$ for each \elem\ $e$.

\noindent{\bf Step 1: Rewriting $Z$.} 
For any uncovered $(\ex)\in A_{k-1}$ in the decision tree ALG at time $\alpha 2^{k-1}$, let $P_\ex$ be the path traced by $(\ex)$ in ALG, starting from time $\alpha 2^{k-1}$ and ending at time $\alpha 2^k$ or when $(\ex)$ is covered.
% By abuse of notations, we will also view $P_\ex$ as the subset of {\it \elem s} selected along this path.

Recall that in the definition of $Z$, for each time $t$ between $\alpha 2^{k-1}$ and $\alpha 2^k$, we sum over all states  $(E,H')$ at time $t$.
Since $t\geq \alpha 2^{k-1}$, and the subset of uncovered scenarios only shrinks at $t$ increases, for any $(E,H')\in V(t)$ we have $H'\sse A_{k-1}$. 
%this means all expanded \sce s in $H'$ must be in $A_{k-1}$, \iow, 
So, only the expanded \sce s in $A_{k-1}$ contribute to $Z$.
Thus we may rewrite~\eqref{eq:gain_*} as
{\small \begin{align}
Z \hspace{3mm}  \,\,&= \,\, \sum\limits_{(\ex)\in A_{k-1}} \pi_\ex \cdot \enskip\sum\limits_{(e;E,H')\in P_\ex} \left(\frac{f_{\ex}(e\cup E)-f_{\ex}(E)}{1-f_{\ex}(E)}+\mathbf{1}[(\ex)\in R_e(H')]\right) \notag \\
&\,\,\le\,\,  \sum\limits_{(\ex)\in A_{k-1}} \pi_\ex \cdot \enskip \left(\sum\limits_{(e;E,H')\in P_\ex} \frac{f_{\ex}(e\cup E)-f_{\ex}(E)}{1-f_{\ex}(E)} \,\,+\,\, \sum\limits_{(e;E,H')\in P_\ex} \mathbf{1}[(\ex)\in R_e(H')]\right). \label{eq:gain-UB}
\end{align}}

\noindent{\bf Step 2: Bounding the Inner Summation.}
The rest of our proof involves upper bounding each of the two terms in the summation over $e\in P_\ex$ for any fixed $(\ex)\in A_{k-1}$.
To bound the first term, we need the following standard result on \submod\ \func s.

\begin{lemma}[\cite{azar2011ranking}] \label{cl:AG-eps}
Let  $f:2^U\rightarrow [0,1]$ be  any monotone function with $f(\emptyset)=0$ and $\eps=\min\{ f(S\cup \{e\}) - f(S) \,: \, e\in U, S\sse U, f(S\cup \{e\}) - f(S) >0 \}$ be the separability \pmt.
Then for any nested sequence of subsets $\emptyset = S_0\sse S_1\sse \cdots S_k\sse U$, it holds \[\sum_{t=1}^k \frac{f(S_t)-f(S_{t-1})}{1-f(S_{t-1})}\,\,\le \,\, 1+\ln\frac{1}{\eps}.\]
\end{lemma}

%As $\eps = \frac 1{m-1}$ for the family $\{f_\ex:(\ex)\in \Omg\}$, 
It follows immediately that 
\begin{equation}
    \label{eq:asr-app2}
\sum_{(e;E,H')\in P_\ex}\frac{f_{\ex}(e\cup E)-f_{\ex}(E)}{1-f_{\ex}(E)}  \,\, \leq  \,\,1 + \ln{\frac{1}{\eps}}.
\end{equation}
%\vnote{Need to change..}

Next we consider the second term $\sum\limits_{(e;E,H')\in P_\ex} \mathbf{1}[(\ex)\in R_e(H')]$. 
%In contrast to the first term, To establish the above \ineq\ we will carefully exploit sparsity of noise and show  that
% \begin{equation}
% \label{eq:num-light} \sum_{e \in P_\ex}{\mathbf{1}[(\ex)\in R_e(H')]}\leq r+\log_{|\Omg|} m\leq r+\log_2 m.
% \end{equation}
Recall that $S\sse [m]$ is  the subset of original \sce s with at least one expanded \sce\ in $H'$. Consider the partition of scenarios $S$ into $|\Omega|+1$ parts based on the response entries (from $\Omega\cup\{*\}$) for element $e$. 
From Algorithm~\ref{alg:asr_*}, recall that $U_e(S)$ denotes the part with response $*$ 
and $C_e(S)$ denotes the largest cardinality part among the non-$*$ responses.  Also, 
$o_e(S)\in \Omega$ is the outcome corresponding to part $C_e(S)$. Moreover, $R_e(H')\sse H'$ consists of all expanded-scenarios that {\em do not} have outcome $o_e(S)$ on element $e$. Suppose that $(\ex)\in R_e(H')$. Then, it must be that the observed outcome on $e$ is {\em not}  $o_e(S)$. Let $S'\sse S$ denote the subset of original scenarios that are also compatible with the observed outcome on $e$. We now claim that $|S'|\le \frac{|S|+r}{2}$. To see this, let $D_e(S)\sse S$ denote the part having the  {\em second largest cardinality} among the non-$*$ responses for $e$. As the observed outcome is not  $o_e(S)$ (which corresponds to the largest part), we have 
$$|S'| \le |U_e(S)| + |D_e(S)| \le |U_e(S)| + \left( \frac{|S|-|U_e(S)|}2 \right) = \frac{|S|+|U_e(S)|}2\le \frac{|S|+r}{2}.$$ 
The first inequality above uses the fact that $S'$ consists of $U_e(S)$ (scenarios with $*$ response) and some part (other than $C_e(S)$) with a non-$*$ response. The second inequality uses $|D_e(S)|\le \frac{|D_e(S)| + |C_e(S)|}{2}\le \frac{|S| - |U_e(S)|}{2}$. The last inequality uses the upper-bound $r$ on the number of $*$ responses per element. It follows that each time $(\ex)\in R_e(H')$, the number of compatible (original) scenarios on path $P_\ex$ changes as $|S'|\le \frac{|S|+r}{2}$. Hence, after $\log_2 m$ such events, the number of compatible scenarios on  path $P_\ex$ is at most $r$. Finally, we use the fact that the number of compatible scenarios reduces by at least one whenever $(\ex)\in R_e(H')$, to obtain 
\begin{equation}
    \label{eq:asr-app3}
\sum\limits_{(e;E,H')\in P_\ex} \mathbf{1}[(\ex)\in R_e(H')]\le r+ \log_2 m.\end{equation}
Combining \eqref{eq:gain-UB}, \eqref{eq:asr-app2} and \eqref{eq:asr-app3}, we obtain the lemma. 
 \qed
 
%%%%%%%%%%%%%%%%% dropped old proof - errors %%%%%%%%%%%%%%%%%%% 
\ignore{
and
$\ell(e; S)$ be the lightest cardinality set among
\[\{i\in S: r_i(e) = o\}, \quad \forall o\in \Omg,\]
and we will call the \corres\ node a {\it light} node.
We then defined \[R_e(H') = \{(\ex)\in H': i \in \ell(e; S)\} = H'\cap \left(\cup_{i \in \ell(e; S)} H_i \right)\]
As the key step, by applying sparsity of noise, we show that if a path turns to a light node for $\log m$ times, then the number of consistent original \hypos\ will drop below $r$.

\begin{lemma}\label{lem:oct5}
Let $v$ be a node in the \dec\ tree ALG and denote the consistent original \hypo s in $v$ as $S_v$.
\Sps\ the path from the root to $v$ contains at least $\log_{|\Omg|} m$ light nodes, then $|S_v|\leq r$.
\end{lemma}

To complete the proof, consider the portion $P'_\ex$ (resp. $P''_\ex$) of path $P_\ex$ before (resp. after) $|S|$ drops below $r$.
Then, Lemma~\ref{lem:oct5} immediately implies that
\[\sum_{(e;E,H') \in P'_{\ex}} \mathbf{1}[(\ex)\in R_e(H')]\leq \log_{|\Omg|} m.\]
To bound the length of $P''_\ex$, note that in each node $(e;E,H')$ on $P''_\ex$, each time $(\ex)\in R_e(H')$, $|S|$ will reduce by at least one. 
Hence, 
\[\sum_{(e;E,H') \in P'_{\ex}} \mathbf{1}[(\ex)\in R_e(H')]\leq r.\]
% \OTOH, by a similar argument we have
Combining the above two \ineqs, we immediately obtain \eqref{eq:num-light}, and hence completes the proof of Lemma~\ref{lem:GUB}.
\qed

\noindent{\bf Proof of Lemma~\ref{lem:oct5}.}
\Sps\ $\ell(e;S) = S^{\bar o}$ for some outcome $\bar o\in \Omg$, and we will call the node following $\bar o$ a {\it light} node.
Thus, for each $(\ex)\in R_e(H')$, we have $i\in \ell(e;S)$. 
Write $S^o := T^o(e) \cap S$ for each $o\in \Omg$ and $S^* = T^*(e)\cap S$.

We first characterize the surviving {\it original} \sce s
in the $\bar o$-branch of the current node $(e;E,H')$ in the \dec\ tree ALG.
If $i\in \ell(e;S)$, then by definition of $\ell(e;S)$ 
the path $P_\ex$ follows the $\bar o$-branch.
If $i\in S^*$, then the expanded \sce s of $i$ in $H'$ will be split evenly across the $|\Omg|$ branches, and hence $i$
will appear in {\it every} branch, in \parti, the $\bar o$-branch. 

Hence, the number of surviving original \sce s in the $\bar o$-branch
is 
\[|\ell(e;S)| + |S^*|
= \left(\frac{|\ell(e;S)|}{|\Omg|} + \sum_{o\neq \bar o} \frac{|S^o|}{|\Omg|}\right) + |S^*| 
\leq \frac{|S|}{|\Omg|} + \left(1-\frac 1{|\Omg|}\right)|S^*| 
\le \frac{|S|}{|\Omg|} + \left(1-\frac 1{|\Omg|}\right) r,\]
where the equality follows since 
$\ell(e;S) = S^{\bar o}$, 
first inequality follows from the definition of $\ell(e;S)$, and the last \ineq\ uses the key fact that $|S^*|\leq r$.
In words, this means each time that the path $P_\ex$ turns to the child \corres\ to $R_e(H')$, the number $s_{new}$ of surviving original \sce s after selecting $e$ is bounded by 
\[s_{new} \le \frac 1{|\Omg|} |S| + \left(1-\frac1{|\Omg|}\right)r.\] 

Therefore, after following the $\ell(e;S)$ branch for $k$ times, the number $s_k$ of surviving original \sce s is bounded by 
\[s_k \leq \frac m{|\Omg|^k} + r\left(1-\frac 1{|\Omg|} \right)\cdot \left(1+\frac 1{|\Omg|} + \frac 1{|\Omg|^2} +...+ \frac 1{|\Omg|^k})\right)\leq 
\frac m{|\Omg|^k} + r.\]
Thus after $k = \log_{|\Omg|} m$ such events, we have $s_k \le r$. \qed}

\section{ASRN with High Noise: Details of Section~\ref{sec:many_star}}
\label{apdx:many_stars}
% In Section~\ref{sec:many_star} we outlined our \alg\ and sketched a proof.
% In this section, we provide the technical details.
% In \ref{subsec:member}, we formally define the membership oracle, and in \ref{subsec:main} we formally state the main \alg. 
% In \ref{subsec:analysis_many_star}, we state the extra lemmas needed for analyzing the main \alg\ and present a formal proof of Theorem~\ref{thm:sparse}, the main result in the noise-heavy regime.
% Finally in the remaining subsections, we present formal proofs of the lemmas stated in Section~\ref{sec:many_star} and  Subsection~\ref{subsec:main}.

\subsection{Details of the  Membership Oracle: Proof of Lemma \ref{prop:membership_oracle}}
\label{subsec:member}
We first describe how to verify whether a given \hypo\ is the true \hypo\ using $(\log m)$ tests.
Select an \arb\ set $W$ of $4\log m$ deterministic tests for $i$, and let $Y$ be the set of consistent \hypos\ after performing these tests.
Without loss of generality, we assume $i\in T^+$ for all $T\in W$.
There are three cases:
\bitem 
\item {\bf Trivial Case:} if $\bar i \in T^-$ for {\it some} $T\in W$, then we  rule out $i$ when any  test $T$ is performed.
\item {\bf Good Case:} if $\bar i \in T^*$ for more than half of the tests $T$ in $W$, then by Chernoff's \ineq, with high \prb\ we observe at least one ``-'', hence ruling out $i$. 
\item {\bf Bad Case:} if $\bar i\in T^+$ for less than half of the tests in $W$, then we may not be able to ruling out $i$ with a high \prb.  
To overcome this, we then test between $i$ with each \hypo\ in $Y$ by selecting a test where these two \hypos\ have distinct \dtmnstc\ outcomes. 
This test exists due to Assumption \ref{assu:iden}.
\eitem 

At this juncture, we formally define the membership oracle ${\rm Member}(Z)$ in \Alg\ \ref{alg:member}.
Note that Steps \ref{member:z-sep}, \ref{member:repeat-test} and \ref{member:indiv-tests} are well-defined because the ODTN instance is assumed to be identifiable. 
If there is no new test in Step~\ref{member:z-sep} with $T^+\cap Z'\ne \emptyset$ and $T^-\cap Z' \neq \emptyset$, then we must have $|Z'|=1$. 
If there is no new test in Step~\ref{member:repeat-test} with $z\not\in T^*$ then we must have identified $z$ uniquely, i.e. $Y=\emptyset$. 
Finally, in step~\ref{member:indiv-tests}, we use the fact that there are tests that deterministically separate every pair of hypotheses.

\begin{algorithm}
\begin{algorithmic}[1]
\State Initialize: $Z'\lar Z$.
\While {$|Z'| \geq  2$} \quad\quad \% While-loop 1: Finding a suspect -- reducing $|Z'|$ to $1$ 
\State\label{member:z-sep} Choose any new test $T\in \T$ with $T^+\cap Z'\ne \emptyset$ and $T^-\cap Z' \neq \emptyset$, observe outcome $\omg_T\in\{\pm 1\}$.
\State Let $R$ be the set of hypotheses ruled out, i.e. $R = \{j\in [m]: M_{T,j} =-\omg_T \}$. 
\State Let $Z'\leftarrow Z'\backslash R$.
\EndWhile
\State\label{member:uniq}Let $z$ be the unique \hypo\ when the while-loop ends. %\quad\quad  \% Identified a ``suspect''.
\Comment{Identified a ``suspect''.}
\State Initialize $k\leftarrow 0$ and $Y=H $.  
\While{$Y \neq \emptyset$ and $k\leq 4\log m$} \label{member:loop} \Comment{While-loop 2: choose deterministic tests for $z$.}
\State \label{member:repeat-test} Choose any new test $T$ with $M_{T,i}\neq *$ and observe outcome $\omg_T\in \{\pm 1\}$. %suppose that $z\in T^+(e)$. (The case $z\in T^-(e)$ is symmetric.) 
\If{$\omg_T = -M_{T,i}$} \Comment{$i$ ruled out.}
\State Declare ``$\bar{i}\not\in Z$'' and stop. 
\Else
\State Let $R$ be the set of hypotheses ruled out, $Y\leftarrow Y\setminus R$ and $k\leftarrow k+1$. 
\EndIf
\EndWhile

\If{$Y=\emptyset$} \label{member:empty}  
\State Declare ``$\bar{i}=i$'' and terminate.
\Else 
\State \label{item:step3} 
Let $W\sse \T$ denote the tests performed in step~\ref{member:repeat-test} and \Comment{Now consider the``bad'' case.}
\beqn\label{eq:member-set}
    J &= \{j\in Y: M_{T,j}= M_{T,i}\ \text{for at least}\ 2\log m\ \text{tests $T\in W$}\}\\
    &=\{j\in Y:  M_{T,j}=*\ \text{for at most}\ 2\log m\ \text{tests $T\in W$}\}.
\eeqn
\State \label{member:indiv-tests} 
For each $j\in J$, choose a test $T=T(j)\in \T$ with $M_{T,j},M_{T,i}\neq *$ and $M_{T,j}=-M_{T,i}$
\State let $W'\sse \T$ denote the set of these tests.
\EndIf

\If{no tests in $W\cup W'$ rule out $i$}\label{member:check} \Comment{Let $i$ duel with \hypos\ in $J$.}
\State Declare ``$\bar{i} = i$''.
\Else
\State Declare ``$\bar{i}\notin Z$''.
\EndIf
\caption{Member$(Z)$ oracle that  checks if $\bar{i}\in Z$. \label{alg:member}}
\end{algorithmic}
\end{algorithm}

Now, let us show that the membership oracle in \Alg\ \ref{alg:member} has cost $O(|Z| + \log m)$ as stated in Lemma \ref{prop:membership_oracle}.

\noindent{\bf Proof of Lemma \ref{prop:membership_oracle}.}
If $\bar{i} \in Z$ then it is clear that $i=\bar{i}$ in step~\ref{member:uniq} and Member$(Z)$ declares $\bar{i} = i$. 
Now consider the case $\bar{i} \not\in Z$. 
Recall that $i\in Z$ denotes the unique hypothesis that is still compatible in step~\ref{member:uniq}, and that $Y$ denotes the set of compatible hypotheses among $[m]\setminus \{i\}$, so it always contains $\bar{i}$. 
Hence,  $Y\ne \emptyset$ in step~\ref{member:empty}, which implies that $k=4\log m$.
Also recall the definition of set $S$ and $J$ from \eqref{eq:member-set}.
\bitem
\item Case 1.
If $\bar{i} \in J$ then we will identify correctly that $\bar{i}\ne i$ in step~\ref{member:check} as one of the tests in $W'$ (step~\ref{member:indiv-tests})  separates $\bar{i}$ and $i$ deterministically. 
So in this case we will always declare $\bar{i} \notin Z$.
\item Case 2. 
If $\bar{i}\not\in J$, then by definition of $J$, we have $\bar{i}\in T^*$ for at least $2\log m$ tests $T\in W$.
As $i$ has a deterministic outcome for each test in $W$, the probability that all outcomes in $W$ are consistent with $i$ is at most $m^{-2}$. 
So with probability at least $1-m^{-2}$, some test in $W$  must have an outcome (under $\bar{i}$)  inconsistent with $i$, and  based on step~\ref{member:check}, we would declare $\bar{i} \notin Z$.
\eitem
In order to bound the cost, note that the number of tests performed are at most: $|Z|$ in step~\ref{member:z-sep}, $4\log m$ in step~\ref{member:repeat-test} and $|J|\leq |Z|$ in step~\ref{member:indiv-tests}, and the proof follows.
% The key step is to bound $|J|$, for which we use Lemma~\ref{lem:markov}.
% Note that the tests in step~\ref{member:indiv-tests} are only performed if $Y\ne \emptyset$ in step~\ref{member:empty}: so we must have $|W|=k=4\log m$.
% By Proposition~\ref{lem:markov}, we have $J = \{i\in Y : D(i)>|W|/2\}$ as $|W|=4\log m$.
% It now follows that  $|J|\le 2Cm^\alpha$. 
% Hence the total number of tests is $|Z|+4\log m + |J|= O(|Z|+m^\alpha)$. 
\qed

%%%%%%%%%%%%% moved to body %%%%%%%%%%%%%%%%
\ignore{
\subsection{The Main Algorithm}\label{subsec:main}
We are now ready to formally state the overall algorithm as in \Alg~\ref{alg:large-noise}. 
The \alg\ maintains a subset of consistent \hypos, and iteratively computes the greediest test, as formally specified in Step~\ref{step:alg-many-star-2}.
At each $t=2^k$ where $k=1,2,...\log m$, we consider a set $Z$ of $O(m^\alpha)$ hypotheses with the smallest number of $\star$-tests selected. 
Then, we invoke
Member$(Z)$ to check whether the target \hypo\ is in $Z$.
If so, the the \alg\ halts and returns the target, \ow\ it continues choosing greedy tests until the next power-of-two \iter.

Note that the membership oracle is invoked $O(\log m)$ times as the total number $t$ of tests used is always at most $O(m)$.
Using Lemma~\ref{prop:membership_oracle}, it is clear that $\bar{i}$ is identified correctly with probability at least $1-\frac1m$. 
We now analyze the cost. The oracle Member is always invoked on $|Z|=O(m^\alpha)$ hypotheses. Using Lemma~\ref{prop:membership_oracle}, the expected  number of tests due to step~\ref{step:alg-many-star-1} is $O(m^\alpha\log m)$. In the rest of this subsection, we will bound the expected cost due to tests in  step~\ref{step:alg-many-star-2}. %and ignore the cost of performing tests due to step~\ref{step:alg-many-star-1}. 

\subsection{Analysis of \Alg~\ref{alg:large-noise}}\label{subsec:analysis_many_star}
\noindent{\bf Truncated Decision Tree.} 
Let $\mathbb{T}$ denote the decision tree corresponding to our algorithm. 
We only consider tests that correspond to step~\ref{step:alg-many-star-2}.
Recall that $H$ is the set of {\it expanded} hypotheses and that any expanded \hypo\ traces a unique path in $\mathbb{T}$.
For any $(\ex)\in H$, let $P_\ex$ denote this path traced;
so $|P_\ex|$ is the number of tests performed in Step~\ref{step:alg-many-star-2} under $(\ex)$. 
We will work with a truncated decision tree $\overline{\mathbb{T}}$, defined below. 

Fix any expanded \hypo\ $(\ex)\in H$. 
For any $t\ge 1$, let $\theta_\ex(t)$ denote the fraction of the first $t$ tests in $P_\ex$ that are $\star$-tests for hypothesis $i$. 
Recall that $P_\ex$ only contains tests from Step~\ref{step:alg-many-star-2}. 
Let $\rho=2$ and define
\begin{equation}\label{def:trunc}
t_\ex = \max\left\{ t \in \{2^0, 2^1,\cdots, 2^{\log m}\} \,\,:\,\, \theta_\ex (t') \ge \frac 1\rho \mbox{ for all }t'\le t \right\}.
\end{equation}
If $t_\ex > |P_\ex|$ then set $t_\ex = |P_\ex|$. 
The truncated decision tree 
$\overline{\mathbb{T}}$ is the subtree of $\mathbb{T}$ consisting of the first $t_\ex$ tests along path $P_\ex$, for each $(\ex)\in H$. 
By abuse of notation we also use $\theta_i(t)$ to denote the {\it random} fraction of $\star$-tests for $i$, with randomness over $\omg$.

\noindent{\bf Relating costs of $\mathbb{T}$ and  $\overline{\mathbb{T}}$.} 
We now show that the cost of $\overline{\mathbb{T}}$ is at most twice that of $\mathbb T$.
Intuitively, at the next power-of-two step after \iter\ $\tau$, which is within another $\tau$ \iter s, the membership oracle will identify $i$ as the target \hypo\ and \alg\ halts.
Thus, the number of iterations {\it after} \iter\ $\tau$ is at most $\tau$, and hence the expected cost of $\mathbb{T}$ is at most twice that of $\overline{\mathbb{T}}$.
\begin{lemma}\label{lem:abar-to-a}
For each $x\in H$, the number of tests performed $|P_x|\le 2\cdot t_x$. 
Hence, the expected cost of $\mathbb{T}$ is at most twice that of $\overline{\mathbb{T}}$. 
\end{lemma}
}

\subsection{Proof of Proposition \ref{lem:lower-bound}: SSC-based Lower Bound on OPT}
Consider any feasible decision tree $\mathbb {T}$ for the ODTN instance and any hypothesis $i\in [m]$. 
If we condition on $\bar{i}=i$, then $\mathbb {T}$ corresponds to a feasible adaptive policy for ${\rm SSC}(i)$. 
In fact,
\bitem 
\item for any expanded \hypo\  $(\omg,i)\in \Omg(i)$, the tests performed in $\mathbb {T}$ must rule out all the hypotheses $[m]\bs \{i\}$, and 
\item the hypotheses ruled-out by any test $T$ (conditioned on $\bar{i}=i$) is a random subset that has the same distribution as $S_T(i)$.
\eitem
Formally, let $P_{i,\omg}$ denote the path traced in $\mathbb{T}$ under test outcomes $\omg$, and $|P_{i,\omg}|$ the number of tests performed along this path. 
Recall that $u_i$ is the number of \unk\ tests for $i$, and that the \prb\ of observing outcomes $\omg$ when $\bar{i}=i$ is $2^{-u_i}$, so this policy for $SSC(i)$ has cost $\sum_{(i,\omg)\in \Omg(i)} 2^{-u_i}\cdot |P_{i,\omg}|$. 
Therefore, \[{\rm OPT}_{{\rm SSC}(i)} \le \sum_{(i,\omg)\in \Omg(i)} 2^{-u_i}\cdot |P_{i,\omg}|.\]
Taking expectations over $i\in [m]$ the lemma follows.
\qed

\subsection{Proof of Lemma~\ref{prop-greedy-star}: Greedy Is Good for Most Hypotheses}
For simplicity, write $(T')^+$ as $T'_+$ (similarly define $T'_-,T'_*$).
Note that $\E[|S_T(i)\cap (A\bs i)|] = \frac12\left( |T^+\cap A| + |T^-\cap A|\right) $ because $i\in T^*$. 
We consider two cases for the test $T'\in \T$.
\begin{itemize}
\item If $M_{T',i}=\star$, then by the definition of the greedy rule (Step \ref{step:alg-many-star-2}), we have 
$$\E[|S_{T'}(i)\cap (A\bs i)|] = \frac12\left( |T'_+\cap A| + |T'_-\cap A|\right) \le \frac12\left(  |T^+\cap A| + |T^-\cap A| \right).$$
\item If $i\in T'_+\cup T'_-$, then 
$$\E[|S_{T'}(i)\cap (A\bs i)|] \le \max\{  |T'_+\cap A|, |T'_-\cap A| \} \le  |T'_+\cap A| + |T'_-\cap A|,$$ which is at most $ |T^+\cap A| + |T^-\cap A|$ by the choice of $T$.
\end{itemize}
Therefore, in either case, the claim holds.\qed

\subsection{Proof of Proposition~\ref{lem:sparse-OPT}: Sparsity-based Lower Bound on OPT}
To ensure the output is correct w.p. $1$, we need to eliminate all $(m-1)$ hypotheses except the true \hypo.
By the definition of $\alpha$-sparse instances, each test eliminates only $O(m^\alpha)$ \hypos, we need to perform $\Omg(\frac{m-1}{m^\alpha}) = \Omega(m^{1-\alpha})$ tests. \qed

\end{document}